\pdfoutput=1

\documentclass[11pt,,table,xcdraw]{article}

\usepackage[preprint]{acl}

\usepackage{times}
\usepackage{latexsym}

\usepackage[T1]{fontenc}

\usepackage[utf8]{inputenc}

\usepackage{microtype}

\usepackage{inconsolata}

\usepackage{graphicx}

\usepackage{times}
\usepackage{epsfig}
\usepackage{graphicx}
\usepackage{amsmath}
\usepackage{amssymb}

\usepackage{multicol}
\usepackage{booktabs}
\usepackage{multirow}
\usepackage{algorithm}
\usepackage{algorithmic}
\usepackage{amsmath}
\usepackage{enumitem}
\usepackage{subcaption} 
\usepackage{caption}    
\usepackage{tcolorbox}
\usepackage{listings}
\usepackage{titletoc}

\newtheorem{theorem}{Theorem}[section]
\newtheorem{proof}{Proof}[section]
\newtheorem{definition}{Definition}[section]

\newtheorem{assumption}{Assumption}[section]

\title{SafePath: Conformal Prediction for Safe LLM-Based Autonomous Navigation}

\author{
  \textbf{Achref Doula\textsuperscript{1}},
  \textbf{Max Mühlhäuser\textsuperscript{1}},
  \textbf{Alejandro Sanchez Guinea\textsuperscript{2}},\\
  \textsuperscript{1}Technical University of Darmstadt, Germany,
  \textsuperscript{2}NTT Data, Luxembourg,\\
  \small{
    \texttt{doula@tk.tu-darmstadt.de, max@informatik.tu-darmstadt.de, alejandro.guinea@global.ntt}
  }
}

\begin{document}
\maketitle

\begin{abstract}

Large Language Models (LLMs) show growing promise in autonomous driving by reasoning over complex traffic scenarios to generate path plans. However, their tendencies toward overconfidence, and hallucinations raise critical safety concerns. We introduce SafePath, a modular framework that augments LLM-based path planning with formal safety guarantees using conformal prediction.
SafePath operates in three stages. In the first stage, we use an LLM that generates a set of diverse candidate paths, exploring possible trajectories based on agent behaviors and environmental cues. In the second stage, SafePath filters out high-risk trajectories while guaranteeing that at least one safe option is included with a user-defined probability, through a multiple-choice question-answering formulation that integrates conformal prediction. In the final stage, our approach selects the path with the lowest expected collision risk when uncertainty is low or delegates control to a human when uncertainty is high. We theoretically prove that SafePath guarantees a safe trajectory with a user-defined probability, and we show how its human delegation rate can be tuned to balance autonomy and safety. Extensive experiments on nuScenes and Highway-env show that SafePath reduces planning uncertainty by 77\% and collision rates by up to 70\%, demonstrating effectiveness in making LLM-driven path planning more safer.
\end{abstract}
 
\section{Introduction}

Safety is a fundamental requirement for autonomous vehicles (AVs) operating in complex urban environments, where uncertainty arising from the unpredictable behavior of pedestrians, cyclists, and other vehicles, combined with dynamic interactions, poses significant challenges. In this context, safety refers to the ability of a planning approach to minimize the likelihood of collisions or other harmful events to an acceptable low threshold, thus preventing catastrophic consequences. Achieving such a safe performance is challenging in dynamic urban environments due to their complexity. 
Large Language Models (LLMs) have recently emerged as powerful tools for generating navigation strategies in autonomous driving, thanks to their ``reasoning'' capabilities over complex traffic scenarios~\cite{llmdrive1,llmdrive2,DriveVLM}. However, their tendency to display overconfidence, produce hallucinated outputs, and generate unreliable uncertainty estimates poses significant safety risks~\cite{hallucin1,hallucin2}. Such vulnerabilities are especially problematic in real-world domains, where even minor errors can lead to severe consequences.

To ensure the reliable use of LLMs
for path planning in autonomous driving, we propose \emph{SafePath}, the first approach, to the best of our knowledge, to integrate LLM-based path generation with conformal prediction to provide formal safety guarantees on the generated paths.
SafePath consists of three stages. First, the \textbf{Path Generation} stage uses an LLM to produce a set of candidate paths by reasoning over the AV goals and environmental cues, such as traffic rules and agent behaviors. Second, in the \textbf{Uncertainty-Aware Path Selection} stage, we refine the set of candidate paths and filter out risky paths. To achieve this, we introduce a multiple choice question-answering (MCQA) procedure that leverages a second LLM and conformal prediction (CP)~\cite{cp0}, to estimate uncertainty and guarantee that the refined set of paths includes, with a user-defined probability, at least one path that meets safety requirements. 

Third, the \textbf{Path Decision} stage determines which path to execute by examining how similar the paths in the refined set from the second stage are. If the set contains a single path, indicating low uncertainty, the path is adopted directly. If the set contains multiple paths that exhibit high similarity, SafePath selects the one with the lowest expected collision risk. Only when paths diverge significantly, indicating high uncertainty, SafePath delegates control to a human to ensure safety.

We provide a theoretical analysis proving that SafePath guarantees, with user-defined probability, that the selected trajectory is safe. To empirically validate our approach, we conduct a comprehensive evaluation across two complementary settings: nuScenes~\cite{nuscenes}, which supports open-loop planning from rich, annotated urban scenes, and highway-env~\cite{hwenv}, which enables closed-loop testing with real-time agent interaction. This combination allows us to assess SafePath's performance in both static reasoning and dynamic control regimes.
The results of our evaluation show that SafePath, under fully autonomous settings, achieves a reduction in planning uncertainty of up to 77\% and a drop in collision rates by up to 70\% compared to state-of-the-art planning methods.

\section{Related work}
\noindent \textbf{Motion Planning for Autonomous Navigation.}
Motion planning is a central challenge in autonomous driving, requiring predictive navigation in dynamic, often congested environments. Early solutions predominantly relied on rule-based heuristics that maintained strict adherence to lane-keeping, safety distances, and traffic regulations~\cite{rb1, rb2, rb3, rb4, rb5}. While interpretable, these systems struggled to accommodate unexpected or extreme scenarios, limiting their robustness~\cite{rb6}. More recent data-driven approaches, including imitation learning~\cite{imit1, imit2, imit3} and reinforcement learning~\cite{rl1, rl2, rl3, rl4}, leverage large-scale datasets to learn adaptive policies capable of responding to evolving conditions. However, their black-box character can impede transparency and trust, which is critical in safety-sensitive domains~\cite{bbox}.

Large Language Models (LLMs) introduce a further dimension of explainability, providing rationale-based explanations of prospective maneuvers~\cite{gptdriver, DriveVLM, llmdrive1, llmdrive2, asyncdriver}. However, LLMs may occasionally hallucinate or exhibit unwarranted overconfidence~\cite{hallucin2, hallucin1}, jeopardizing safety if not adequately constrained. The approach proposed in this work combines LLMs with conformal prediction, preserving the explainability gains of language-based planning while mitigating overconfident or incorrect outputs that could lead to unsafe actions.




\noindent \textbf{Uncertainty Management in Autonomous Driving.}
Urban driving involves uncertainties caused by changing traffic conditions, sensor noise, and incomplete knowledge of other agent intentions~\cite{lvis1, lvis2, lvis3}. Probabilistic approaches based on Bayesian inference~\cite{bayes1, bayes2, bayes3} and stochastic formulations, including POMDPs~\cite{pomdp}, model these uncertainties but can be computationally heavy and rely on assumptions about underlying data distributions. Redundancy and ensemble-based strategies~\cite{ensb1}, as well as uncertainty-aware reinforcement learning~\cite{url1, url2, url3, rl4}, offer additional means of handling unpredictable environments but often lack formal statistical guarantees.

Conformal prediction avoids strict distributional assumptions and provides statistically valid confidence bounds on its predictions~\cite{ucp1, cp2}, making it promising for safety-critical planning. Recent efforts to integrate conformal prediction with LLMs~\cite{knowno, cpllm1, cpllm2, cpllm3, sun2024conformal} show its potential for improving reliability. In this work, We build on these insights by leveraging conformal prediction to mitigate LLM overconfidence, yielding motion plans with explicit safety guarantees in the face of uncertainty.




\begin{figure*}
    \centering
    \includegraphics[width=\linewidth]{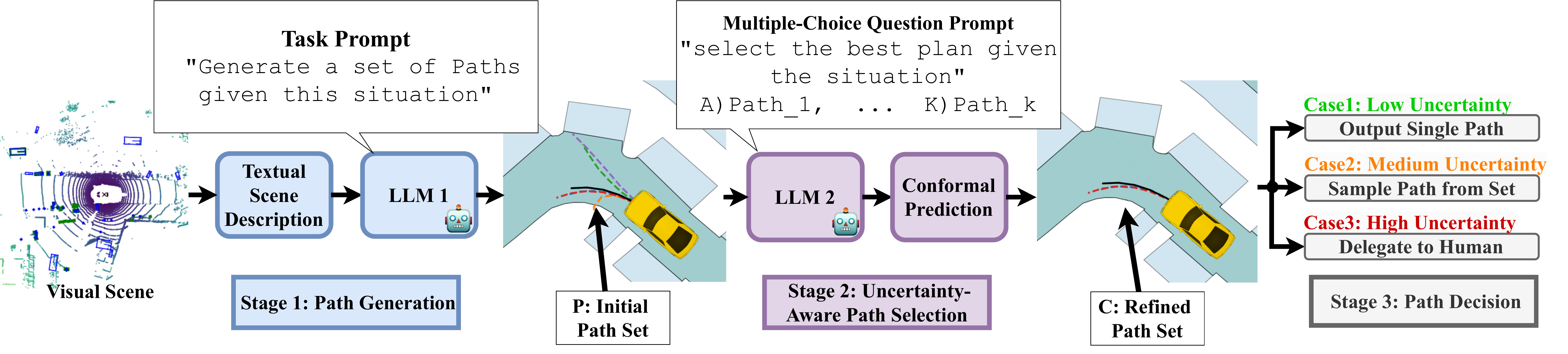} 
    \caption{Overview of SafePath for autonomous vehicle path planning with LLMs. The approach consists of three stages: (1) \textbf{Path Generation}, where $\text{LLM}_1$ generates a set of candidate paths $P$ based on the autonomous vehicle's goals and environmental state derived from a textual scene description; (2) \textbf{Uncertainty-Aware Path Selection}, where $\text{LLM}_2$ refines the paths in $P$ by framing the problem as a multiple-choice question-answering task and performing a conformal prediction step to construct a prediction set $C$ that includes a safe path with user-defined probability $1-\alpha$; and (3) \textbf{Path Decision}, which evaluates the selected path(s) based on uncertainty levels to determine if the vehicle can proceed autonomously by selecting the path with the least expected collision risk or if human intervention is required due to high uncertainty. The prompts shown in the Figure (Task Prompt, Multiple-Choice Question Formulation Prompt) are partial examples for illustration. The complete prompts are provided in Appendix~\ref{app:prompt}.}
    \label{fig:system}
\end{figure*}

\section{Background on Conformal Prediction}\label{sec:background_cp}

Conformal prediction is a statistical framework that transforms the outputs of a deep learning model into prediction sets guaranteed to contain the true outcome with a probability $1-\alpha$, where $\alpha$ is user-defined error rate~\cite{cp1,cp2}. Unlike point estimates, the prediction sets explicitly quantify uncertainty by including all plausible candidates, making CP ideal for safety-critical systems where overconfidence must be mitigated. While we later apply CP to the selection of safe paths in autonomous navigation, we first introduce its general form here.

\noindent\textbf{Constructing Conformal Prediction Sets.}
Let $\mathcal{X}$ be the input space and $\mathcal{Y}$ the label space. Suppose we have a trained predictive model $f_{\theta}$ that provides (possibly overconfident) scores over $\mathcal{Y}$. Conformal prediction seeks to construct, for any new input $x_{\text{test}} \in \mathcal{X}$, a \emph{prediction set} $C(x_{\text{test}}) \subseteq \mathcal{Y}$ such that the true label $y_{\text{test}}$ is included with probability at least $1 - \alpha$. Formally,
\begin{equation}\label{eq:marginal_coverage}
    \mathbb{P}\bigl[y_{\text{test}} \in C(x_{\text{test}})\bigr] \,\ge\, 1 - \alpha,
\end{equation}
where $\alpha$ is a user-specified error rate.
Two key components underlie CP:

\begin{itemize}[leftmargin=*]
    \item \emph{Calibration Set.} A dataset $D_{\text{cal}} = \{(x_i, y_i)\}_{i=1}^n$ of \emph{exchangeable} samples, drawn from the same distribution as the training and test data.\\
    
    \item \emph{Non-Conformity Score.} A function $S(x,y)$ that measures how ``atypical'' a label $y$ is for the input $x$. Examples include $1 - \max(\mathrm{model\;probabilities})$ or model-specific loss values.\\
\end{itemize}
By computing $S(x_i, y_i)$ for each $(x_i, y_i) \in D_{\text{cal}}$, we obtain an empirical distribution of scores. The $(1 - \alpha)$-quantile of these scores, denoted $\hat{q}$, then serves as a threshold for constructing prediction sets at test time.

\noindent\textbf{Constructing Prediction Sets.}
Given a new input $x_{\text{test}}$ with unknown $y_{\text{test}}$, the \emph{prediction set} is:
\begin{equation}\label{eq:cp_prediction_set}
    C(x_{\text{test}}) \;=\; \bigl\{\, y \in \mathcal{Y} : S(x_{\text{test}}, y) \,\le\, \hat{q}\bigr\}.
\end{equation}
A fundamental result, stated in Theorem~\ref{th:cp} ensures that under mild assumptions, specifically, exchangeability in $D_{\text{cal}}$ and the new sample $x_{\text{test}}$, this construction satisfies the coverage criterion in Equation~\ref{eq:marginal_coverage}.

\begin{theorem}[Conformal Prediction]\label{th:cp}
Let $D_{\text{cal}} = \{(x_i, y_i)\}_{i=1}^n$ be a calibration set of exchangeable samples, and let $S$ be a chosen non-conformity score. For each $(x_i, y_i)\in D_{\text{cal}}$, compute $S(x_i, y_i)$ and let $\hat{q}$ be the $(1-\alpha)$-quantile of these scores. Then for a new input $x_{\text{test}}$,
\begin{equation}
   C(x_{\text{test}}) \;=\; \bigl\{\,y \in \mathcal{Y} : S(x_{\text{test}}, y)\,\le\,\hat{q}\bigr\}
\end{equation}
satisfies $\mathbb{P}\bigl[y_{\text{test}} \in C(x_{\text{test}})\bigr] \,\ge\, 1 - \alpha.$
\end{theorem}

\section{Approach}

\subsection{Approach Overview}
\emph{SafePath} ensures the safe use of LLMs in autonomous path planning by providing formal safety guarantees on the output of the LLM-generated paths. 
As illustrated in Figure~\ref{fig:system}, SafePath consists of three sequential stages: Path Generation, Uncertainty-Aware Path Selection, and Path Decision.

In the \textbf{Path Generation} stage, the AV employs a first language model, $LLM_1$, to generate a diverse set of candidate paths based on its mission goals and the current environmental state, which includes perception data and predicted movements of surrounding agents. Each candidate path is accompanied by a rationale, enhancing interpretability and transparency in the decision-making process.

The \textbf{Uncertainty-Aware Path Selection} stage refines the generated set of paths to address the inherent variability and uncertainty in LLM outputs. This refinement process employs a second LLM, $LLM_2$, to select paths from the generated candidates. We frame this selection as a multiple choice question-answering task, where $LLM_2$ receives the relevant scenario context and chooses the most suitable path among the candidates. The structured and bounded MCQA formulation enables us to apply conformal prediction to calibrate the selection process and account for uncertainty. After conformal prediction, the output of this stage is a prediction set of paths that is guaranteed to include at least one path that meets the required safety criteria, with a user-defined probability.

In the \textbf{Path Decision} stage, the autonomous vehicle (AV) evaluates the prediction set obtained from the previous stage to determine the final decision for execution. If the set contains a single path, the AV adopts it immediately. If the set contains multiple paths that are sufficiently similar—indicating a consensus among safe trajectories—the AV will autonomously proceed using the path deemed to have the lowest expected collision risk. In scenarios where the set contains multiple diverging paths, reflecting a high degree of uncertainty, the system defers to human oversight to resolve the ambiguity.

\subsection{Path Generation}

In the Path Generation stage, the AV employs a fine-tuned LLM that utilizes chain-of-thought reasoning (CoT)~\cite{cot} to generate a set of candidate paths based on its current state, perception data, and mission goals. Let \( x = \{x_{\text{percept}}, x_{\text{pred}}, x_{\text{goal}}\} \) represent the AV's perception data, predicted movements of other agents, and mission objectives. By leveraging chain-of-thought reasoning, the LLM explores various possible trajectories, considering the AV's goals and the behavior of surrounding agents.

Formally, the LLM generates a set of $k$ candidate paths \( P = \{p_1, p_2, \dots, p_k\} \) and their associated rationales \( R = \{r_1, r_2, \dots, r_k\} \) through chain-of-thought reasoning:
\begin{equation}
\{(p_i, r_i)\}_{i=1}^{k} = \text{LLM}_1(x)
\end{equation}
Each path \( p_i \) represents a potential trajectory for the AV, while \( r_i \) provides the detailed reasoning process that led to the selection of that path. The CoT rationale enhances transparency and can be used for further analysis or to explain decisions to human operators. Furthermore, several works, such as~\cite{llmdrive1, llmdrive2, cot}, have demonstrated that generating rationales alongside the output leads to more reliable and interpretable results. We provide the full text of the prompts we use to generate the candidate path in Appendix~\ref{app:prompt}.

\subsection{Uncertainty-Aware Path Selection}
\label{sec:uncertainty_path_selection}
Despite fine-tuning, $LLM_1$, may generate unsafe paths due to overconfident or hallucinated outputs~\cite{fine-hal}.
To address this limitation, SafePath integrates conformal prediction to provide statistical guarantees that the selected path set contains at least one \emph{safe} trajectory with probability $1- \alpha$. However, a critical challenge arises from the unstructured nature of LLM-generated paths, which vary in semantic detail and lack a predefined hypothesis space, both essential for conformal prediction guarantees. We first formalize what we mean by a safe path:

\begin{definition}[Safe Path]\label{def:safepath}
A path $p$ is considered \emph{safe} if it satisfies the following constraints:

\noindent\textbf{Collision Avoidance:} $p$ ensures no intersections with other agents or static obstacles in the environment.

\noindent\textbf{Rule Compliance:} $p$ adheres to all traffic laws and operational constraints.

\noindent\textbf{Safety Margins:} $p$ maintains a safe distance from other agents, given their predicted trajectories.
\end{definition}

\noindent Let $\mathcal{S}_x$ denote the set of safe paths given a scenario $x$. For our calibrations and guarantees to hold, we assume:

\begin{assumption}\label{assum:safepath}
The driving datasets used to train and calibrate the planning models contain only safe paths.
\end{assumption}
Assumption~\ref{assum:safepath} ensures that the models and calibration processes are based on safe driving behaviors, which is reasonable given that training data for autonomous vehicles are curated to reflect correct and safe operations and used to train models that do so.

To reconcile unstructured LLM outputs with conformal prediction’s requirements, we reformulate path selection as a multiple-choice question-answering (MCQA) task.
Each candidate path $p_i \in P$ is assigned a distinct label $l_{p_i}$ (e.g., ``A,'' ``B,'' ``C''). Then we prompt a second LLM, $LLM_2$ to select a single letter corresponding to the most suitable path $l^*$, given the scenario description $x$, the set of paths $P$, and their rationale $R$:
\begin{equation}
  l^* = LLM_2(P, R, x).
\end{equation}
Casting path selection as a multiple choice task restricts the response space of $LLM_2$ to a finite set of options, which enables the computation of logit-based probabilities even for closed-source LLMs through their APIs. By extracting the logits, it is possible to compute non-conformity scores to construct statistically valid prediction sets.
Under Assumption~\ref{assum:safepath}, we now define a non-conformity score $s_{p_i}$ for each path $p_i$ using the Least Ambiguous Classification (LAC) score~\cite{lac}\footnote{It is possible to use other non-conformity scores such as RAPS~\cite{raps}, and APS~\cite{aps}. However, as we show in Section~\ref{sec:cp-eval}, LAC yielded the best results.}:
\begin{equation}
  s_{p_i} \;=\; 1 - \mathrm{Softmax}\bigl(\mathrm{logit}(l_{p_i})\bigr),
\end{equation}
where $\mathrm{logit}(l_{p_i})$ is the logit corresponding to the label $l_{p_i}$. Using a held-out calibration set $\mathcal{D}_{\text{cal}}=\{(x_i, p_i)\}_{i=1}^n$, we compute $s_{p_i}$ for each calibration point, then let $\hat{q}$ be the $(1-\alpha)$ quantile of these scores:
\begin{equation}
    \hat{q} = \mathrm{Quantile}\Bigl(\bigl\{s_i\bigr\}_{i=1}^n,\,\frac{\lceil(n+1)(1-\alpha)\rceil}{n}\Bigr),
\end{equation}
where $\lceil\cdot\rceil$ is the ceiling function. For a new situation $x_{\text{test}}$, we compute the non-conformity score $s_{p_i}$ for every candidate path $p_i$, and collect all paths whose scores do not exceed $\hat{q}$:
\[
C \;=\; \{\, p_i \in P : s_{p_i} \,\le\, \hat{q}\}.
\]
This \emph{prediction set} $C$ contains, with probability at least $1 - \alpha$, a correct (and thus safe) path for the new scenario. Concretely, we have:

\begin{theorem}[Prediction Set Safety Guarantee]
\label{th:safepath_1}
Let $C \subseteq P$ be the prediction set generated above at confidence level $1 - \alpha$, and suppose Assumption~\ref{assum:safepath} holds. Then, 
\begin{equation}
\mathbb{P}\bigl[\exists\, p \in C\,\cap\,\mathcal{S}_x\bigr] \;\ge\; 1 - \alpha.
\end{equation}
\end{theorem}

\noindent Theorem~\ref{th:safepath_1} states that the prediction set $\mathcal{C}$ produced by conformal prediction will, with high probability, contain at least one path that meets the safety requirements of Definition~\ref{def:safepath}. We prove Theorem~\ref{th:safepath_1} by leveraging the statistical guarantees of conformal prediction, which ensures that the prediction set includes the true path with probability at least $1 - \alpha$ and datasets are assumed to contain safe paths. The detailed proof is provided in Appendix~\ref{app:proofs}.


\subsection{Path Decision}
\label{sec:pathdec}

Once we have a prediction set $C$ of candidate paths, we determine which path to execute. If $C$ contains a single path, i.e., $|C|=1$, the AV adopts it immediately since it is included with probability $1-\alpha$ of being safe. Otherwise, we evaluate how similar the paths in $C$ are to each other. 

\begin{definition}[$\delta$-Equivalence]\label{def:delta}
Two paths $p_i, p_j \in P$ are \textit{$\delta$-equivalent} if their similarity, as measured by a function $\text{Sim} : P \times P \to [0,1]$, satisfies:
\begin{equation}
\text{Sim}(p_i, p_j) \geq \delta,
\end{equation}
where $\delta \in [0, 1]$ is a predefined threshold. This ensures that paths with small deviations, both in geometry and operational characteristics, are considered interchangeable in terms of safety and functionality\footnote{In Definition~\ref{def:delta}, $\text{Sim}$ is defined as $\text{Sim} : P \times P \to [0,1]$. In Algorithm~\ref{alg:safepath}, we extend this notation to sets, implying that similarity is computed pairwise for all path pairs.} .
\end{definition}
If all pairs of paths in $C$ are $\delta$-equivalent (i.e., the set $C$ effectively contains similar paths), 
the AV selects the path with the lowest expected collision based on predictions in the visual scene description. 
We define $\mathrm{Collision}(p)$ as the fraction of agents with which $p$ intersects out of the total number of agents: 
\begin{equation}
\mathrm{Collision}(p) 
= 
\frac{\#\{\text{agents collided with } p\}}
{\#\{\text{all agents in the scene}\}}.
\end{equation}
Hence, the final path is chosen according to
\begin{equation}
p_{\mathrm{final}} 
= 
\arg\min_{p \in C} \mathrm{Collision}(p).
\end{equation}
If, on the other hand, there exists at least one pair $(p_i, p_j)$ with $\mathrm{Sim}(p_i, p_j) < \delta$, the AV defers to a human operator to avoid committing to highly divergent or conflicting trajectories.

\medskip
\noindent\textbf{SafePath Guarantee.} 
By combining the coverage property from Theorem~\ref{th:safepath_1} with the final decision mechanism, we obtain:

\begin{theorem}[SafePath Guarantee]\label{th:safepath_2}
Let $\mathcal{C}$ be the prediction set generated in the Uncertainty-Aware Path Selection (Section~\ref{sec:uncertainty_path_selection}) at confidence level $1-\alpha$, and suppose 
\[
\mathbb{P}\bigl[\exists\, p \in \mathcal{C}\,\cap\,\mathcal{S}_x\bigr] \;\ge\; 1 - \alpha.
\]
Let $p_{\text{final}}$ be the path chosen according to the path-decision procedure above. Then,
\begin{equation}
\mathbb{P}\bigl[\,p_{\text{final}} \in \mathcal{S}_x\bigr] \;\ge\; 1 - \alpha.
\end{equation}
\end{theorem}

\noindent In other words, once the prediction set is guaranteed to contain a safe path with probability at least $1 - \alpha$, the additional selection or sampling steps (including possible human intervention) preserve that probability. Hence, SafePath produces a single final path that is safe with the same high probability. We provide a full proof of Theorem~\ref{th:safepath_2} in Appendix~\ref{app:proofs}.
The complete SafePath algorithm is shown in Algorithm~\ref{alg:safepath}.

\begin{algorithm}
\caption{SafePath Algorithm}
\label{alg:safepath}
\begin{algorithmic}[1]
\STATE \textbf{Input:} Current state $x = \{x_{\text{percept}}, x_{\text{pred}}, x_{\text{goal}}\}$, perception data, mission goals
\STATE \textbf{Output:} Final path $p_{\text{final}}$

\STATE \textcolor{blue}{\textbf{Stage 1: Path Generation}}
\STATE \quad Use LLM with chain-of-thought reasoning to generate candidate paths $P = \{p_1, p_2, \dots, p_k\}$ and rationales $R = \{r_1, r_2, \dots, r_k\}$
\STATE \quad  $\{(p_i, r_i)\}_{i=1}^{k} = \text{LLM}_1(x)$

\STATE \textcolor{blue}{\textbf{Stage 2: Uncertainty-Aware Path Selection}}
\STATE \quad Frame the selection as a multiple choice question-answering task
\STATE \quad Constrain LLM to output single letters representing each path
\STATE \quad Extract probabilities for each path $p_i$ and compute non-conformity scores for each path $s_{p_i}=1-softmax(logit(l_{p_i}))$
\STATE \quad Construct prediction set $C = \{ p_i \mid s_{p_i} \leq \hat{q} \}$

\STATE \textcolor{blue}{\textbf{Stage 3: Path Decision}}
\IF{$|C| = 0$}
    \STATE \quad  Delegate to human to take over the control.
    
\ELSIF{$|C| = 1$}
    \STATE \quad Adopt the single path in $C$ as $p_{\text{final}}$
\ELSE
    \STATE \quad Compute path similarity
    \IF{$Sim(C) \geq \delta$}
        \STATE \quad \quad $p_{\mathrm{final}} \;=\; \arg\min_{p \in C} \; \mathrm{Collision}(p)$
    \ELSE
        \STATE \quad Delegate to human to take over the control.
    \ENDIF
\ENDIF

\end{algorithmic}
\end{algorithm}

\paragraph{Implications.} 
The usage of the path decision stage reduces the prediction set to a size of 1 which has the advantage of reducing uncertainty and providing the planner with a feasible path, rather than a set. The parameter $\delta$ icontrols how strictly paths must match in similarity: a higher value prompts more human intervention when predictions diverge, preventing overconfident choices, while a lower $\delta$ preserves greater autonomy by allowing the AV to select among less similar paths.

\section{Evaluation}

To empirically validate SafePath, we structure our evaluation around three core goals: (1) understanding how delegation can be adaptively controlled, (2) assessing the statistical validity and efficiency of our conformal prediction layer, and (3) benchmarking overall planning safety in diverse settings.

We begin by examining the effect of the path-similarity threshold $\delta$, which governs human delegation and autonomy. Next, we evaluate SafePath's coverage performance across different confidence levels ($\alpha$) using conformal prediction, and compare it to state-of-the-art conformal methods. Finally, we assess collision-avoidance performance in both open-loop and closed-loop settings: open-loop tests SafePath’s reasoning from static inputs, while closed-loop evaluates SafePath in interactive scenarios where decisions influence future states.

Additional results are in the Appendix, including additional metrics~(App.~\ref{app:l2}), aggregation strategies~(App.~\ref{app:aggreg}), other LLMs~(App.~\ref{app:llms}), and qualitative examples~(App.~\ref{app:qual1}).

\subsection{Dataset and Simulation Environments}
We evaluate SafePath using two complementary testbeds covering distinct planning modes. For open-loop evaluation, we use nuScenes~\cite{nuscenes}, a real-world dataset comprising 1,000 driving scenes, and over 40,000 annotated frames.

For closed-loop evaluation, we use \texttt{highway-env}~\cite{hwenv}, a simulation environment that supports interactive agent control. It includes scenarios such as highways, intersections, and roundabouts, allowing SafePath to be tested in settings where each decision affects future states.

\subsection{Implementation}
We use \texttt{gpt-3.5-turbo-0125} for both path generation (LLM1) and path selection (LLM2). LLM1 is fine-tuned on nuScenes for one epoch (cost: \$120 via OpenAI API), while LLM2 remains frozen. Both LLMs use system prompts with few-shot examples (Appendix~\ref{app:prompt}).

Conformal prediction is calibrated using 5,000 held-out training samples. Path similarity is computed via Sentence-BERT~\cite{sbert} embeddings and cosine distance.
All LLM queries are made via the OpenAI API. The codebase is implemented in Python. Additional results with alternative LLM2s are in Appendix~\ref{app:llms}. Our Code will be released upon publication.


\subsection{Effects of Path Similarity Threshold $\delta$}\label{app:delta}

\begin{figure}
    \centering
    \includegraphics[width=0.9\linewidth]{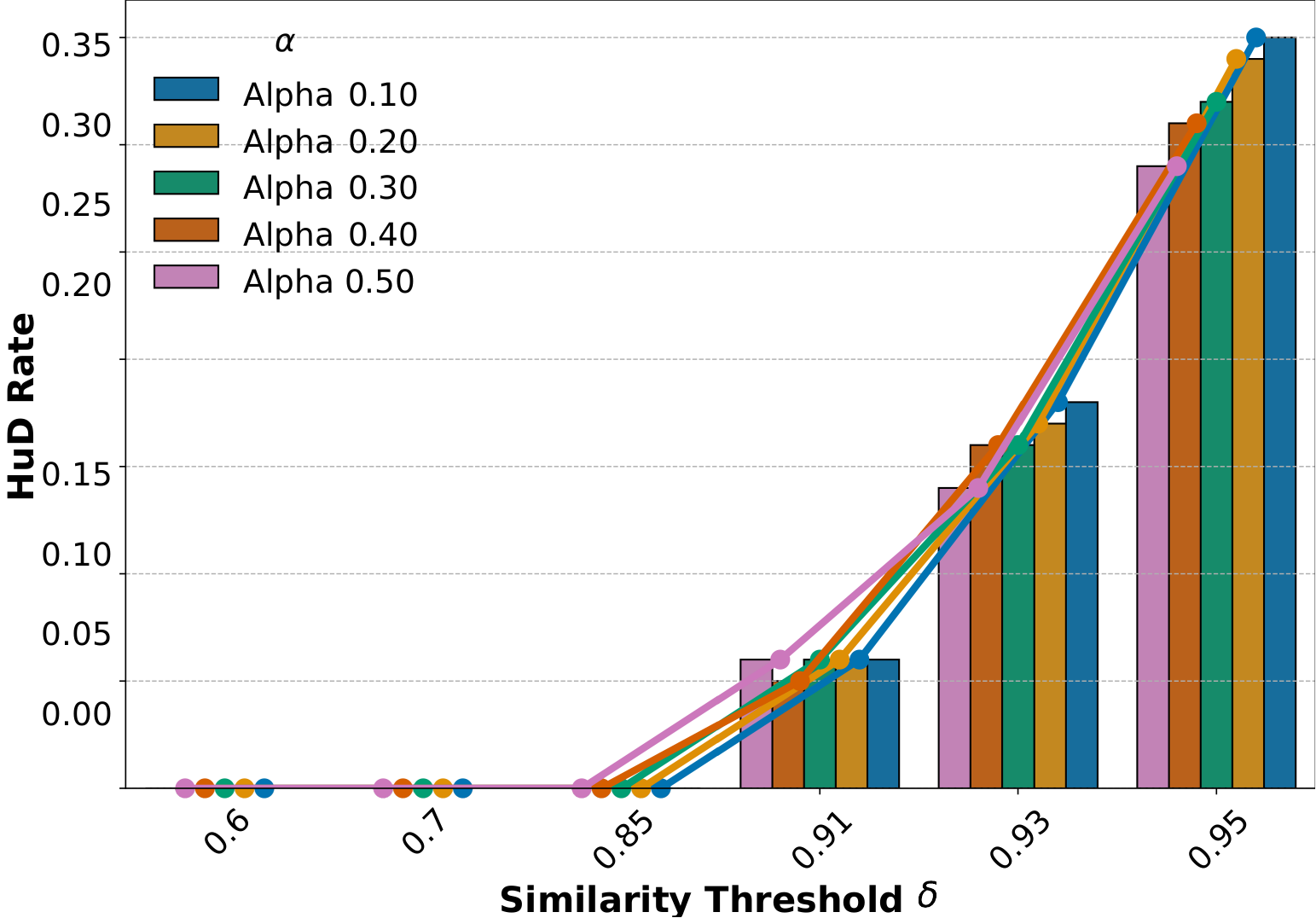}
    \caption{Human delegation rate as a function of the similarity threshold $\delta$ for different $\alpha$ values.}
    \label{fig:delta}
\end{figure}

We analyze how the similarity threshold $\delta$ affects human delegation (HuD). Higher $\delta$ enforces stricter agreement among paths, increasing intervention when candidates differ. Figure~\ref{fig:delta} shows HuD rates across $\alpha \in {0.05, 0.1, 0.2, 0.3}$, where $\alpha$ is the conformal prediction error tolerance.

As $\delta$ increases, SafePath requires greater path similarity, leading to more frequent human intervention. At $\alpha = 0.10$, increasing $\delta$ from $0.85$ to $0.95$ raises HuD from $0.00$ to $0.35$, indicating stricter caution. Yet, HuD remains bounded, and autonomy is preserved at or below $\delta = 0.85$. Notably, $\delta$ has a stronger impact on HuD than $\alpha$; for example, at $\delta = 0.95$, raising $\alpha$ from $0.10$ to $0.30$ only reduces HuD from $0.35$ to $0.32$.

\emph{Unless otherwise stated, we set $\delta = 0.85$ to ensure $\text{HuD} = 0$, enabling fully autonomous operation in all remaining experiments.}

\subsection{Conformal Prediction Evaluation}\label{sec:cp-eval}
We evaluate the conformal prediction step to assess SafePath’s ability to construct prediction sets that reliably include the correct path. For this experiment only, the ground-truth trajectory is added to the candidate set to directly measure coverage. Evaluation is performed across confidence levels ($\alpha$) using $\delta = 0.85$ (no human delegation).

\paragraph{Baselines.}
We compare SafePath against adaptive predictive sets (APS)~\cite{aps}, which guarantees conditional coverage, regularized adaptive predictive sets (RAPS)~\cite{raps}, which aims to reduce set sizes, and ``knowNo''~\cite{knowno}, a planning framework with conformal prediction and human delegation. We run experiments at multiple $\alpha$ values, corresponding to different target coverage rates $1-\alpha$.

\paragraph{Metrics.} In this evaluation we use the following metrics that are proper for conformal prediction. 
\begin{itemize}[leftmargin=*]
    \item \noindent \textbf{Deviation from Target Coverage (DTC)}. The deviation from the target coverage rate (DTC) quantifies the discrepancy between the achieved coverage, \(Cov_{approach}\), and the desired target coverage, \(Cov_{target}=1-\alpha\).

    \item \noindent \textbf{Set Size (SS)}. This metric represents the average set size of the prediction sets. The higher the cardinality of the set, the higher the uncertainty.

    \item \noindent \textbf{Human Delegation Rate (HuD)}. This metric represents the rate of cases that are delegated to humans in case of high uncertainty.
\end{itemize}

\begin{table*}[ht]
\centering
\scalebox{0.75}{
\begin{tabular}{l|ccc|ccc|ccc|ccc|ccc}
\hline
                           & \multicolumn{3}{c|}{$\alpha = 0.1$}                                                                                    & \multicolumn{3}{c|}{$\alpha = 0.2$}                                                                                    & \multicolumn{3}{c|}{$\alpha = 0.3$}                                                                                    & \multicolumn{3}{c|}{$\alpha = 0.4$}                                                                                    & \multicolumn{3}{c}{$\alpha = 0.5$}                                                                                     \\
\multirow{-2}{*}{Approach} & DTC                                    & SS                                    & HuD                                   & DTC                                    & SS                                    & HuD                                   & DTC                                    & SS                                    & HuD                                   & DTC                                    & SS                                    & HuD                                   & DTC                                    & SS                                    & HuD                                   \\ \hline
APS                        & -0.04                                  & 4.42                                  & -                                     & -0.08                                  & 3.99                                  & -                                     & -0.10                                  & 3.60                                  & -                                     & -0.12                                  & 3.13                                  & -                                     & -0.14                                  & 2.62                                  & -                                     \\
RAPS                      & \cellcolor[HTML]{B9FCD0}\textbf{-0.02} & 4.49                                  & -                                     & -0.05                                  & 3.99                                  & -                                     & \cellcolor[HTML]{B9FCD0}\textbf{-0.04} & 3.70                                  & -                                     & -0.11                                  & 3.10                                  & -                                     & -0.14                                  & 2.59                                  & -                                     \\
KnowNo                     & 0.07                                   & 4.70                                  & 1.0                                   & 0.12                                   & 4.25                                  & 1.0                                   & 0.16                                   & 3.59                                  & 1.0                                   & 0.15                                   & 2.88                                  & 1.0                                   & 0.15                                   & 2.24                                  & 1.0                                   \\ \hline
SafePath  ($HuD=0$)  & -0.03                                  & \cellcolor[HTML]{B9FCD0}\textbf{1.00} & \cellcolor[HTML]{B9FCD0}\textbf{0.00} & \cellcolor[HTML]{B9FCD0}\textbf{-0.03} & \cellcolor[HTML]{B9FCD0}\textbf{1.00} & \cellcolor[HTML]{B9FCD0}\textbf{0.00} & -0.05                                  & \cellcolor[HTML]{B9FCD0}\textbf{1.00} & \cellcolor[HTML]{B9FCD0}\textbf{0.00} & \cellcolor[HTML]{B9FCD0}\textbf{-0.08} & \cellcolor[HTML]{B9FCD0}\textbf{1.00} & \cellcolor[HTML]{B9FCD0}\textbf{0.00} & \cellcolor[HTML]{B9FCD0}\textbf{-0.13} & \cellcolor[HTML]{B9FCD0}\textbf{1.00} & \cellcolor[HTML]{B9FCD0}\textbf{0.00} \\ \hline
\end{tabular}}
\caption{Comparison of SafePath, APS~\cite{aps}, RAPS~\cite{raps}, and KnowNo~\cite{knowno} across confidence levels ($\alpha$), using Deviation from Target Coverage (DTC), set size (SS) and Human Delegation Rate (HuD) for KnowNo. All results use $\delta = 0.85$ (HuD = 0). Best results are bolded with light green shading.}

\label{tab:res-cp}
\end{table*}

\paragraph{Results Analysis.} Table~\ref{tab:res-cp} shows that SafePath achieves consistent performance across all confidence levels ($\alpha$). Its path decision stage reduces the effective prediction set size (SS) to 1.00 by selecting the path with the lowest expected collision risk when candidates are sufficiently similar, eliminating the need for human intervention.

In terms of Deviation from Target Coverage (DTC), SafePath remains close to zero and on par with APS and RAPS. At $\alpha = 0.2$, for example, SafePath yields a DTC of $-0.03$, compared to $-0.08$ for APS and $-0.05$ for RAPS, confirming reliable coverage with the specified confidence.

SafePath also has a clear advantage in Human Delegation rate (HuD): with $\delta = 0.85$, it achieves $\text{HuD} = 0.00$ at all $\alpha$ values, ensuring fully autonomous operation. In contrast, KnowNo delegates 100\% of decisions, as it lacks a mechanism to control delegation frequency under uncertainty.

While APS and RAPS maintain low DTC, their prediction sets are larger (SS $\approx$ 2.6--4.5), reflecting higher uncertainty and no internal aggregation strategy. SafePath's $\delta$-equivalence mechanism effectively consolidates paths into a single decision, reducing uncertainty by 77\%.

\begin{table}[ht]
\centering
\scalebox{0.8}{
\begin{tabular}{lcccc}
\hline
\multirow{2}{*}{\textbf{Method}} & \multicolumn{4}{c}{\textbf{Collision (\%) $\downarrow$}} \\
\cline{2-5}
 & \textbf{1s} & \textbf{2s} & \textbf{3s} & \textbf{Avg.} \\
\hline
NMP~\cite{nmp}       & –    & –    & 1.92 & –    \\
SA-NMP~\cite{nmp}    & –    & –    & 1.59 & –    \\
FF~\cite{ff}         & 0.06 & 0.17 & 1.07 & 0.43 \\
UniAD~\cite{uniad}   & 0.05 & 0.17 & 0.71 & 0.31 \\
VAD-Base~\cite{vad}  & 0.07 & 0.17 & 0.41 & 0.22 \\
VLP-VAD~\cite{vlp}  & 0.12 & 0.17 & 0.42 & 0.23 \\
DriveVLM~\cite{DriveVLM}  & 0.10 & 0.22 & 0.45 & 0.27 \\
GPT-Driver~\cite{gpt-driver}  & 0.07 & 0.15 & 1.10 & 0.44 \\
\hline
SafePath (ours) & \cellcolor[HTML]{B9FCD0}\textbf{0.04} & \cellcolor[HTML]{B9FCD0}\textbf{0.12} & \cellcolor[HTML]{B9FCD0}\textbf{0.24} & \cellcolor[HTML]{B9FCD0}\textbf{0.13} \\
\hline
\end{tabular}}
\caption{Collision rate performance on the nuScenes dataset. For a fair comparison, we compare our model with $0\%$ human delegation. The best-performing results are highlighted in bold with light green shading.}
\label{tab:sota_collision}
\end{table}

\subsection{Comparison with State-of-the-Art}
We evaluate SafePath on the nuScenes benchmark, using collision rate as the primary metric. Additional metrics (e.g., L2) are reported in Appendix~\ref{app:l2}.

Table~\ref{tab:sota_collision} shows that SafePath achieves the lowest average collision rate (0.13\%) among all methods, reducing collisions by 52\% compared to DriveVLM (0.27\%) and 70\% compared to GPT-Driver (0.44\%). It also consistently outperforms others across time horizons; at 3 seconds, it achieves 0.24\% collision, 47\% lower than DriveVLM (0.45\%) and far below GPT-Driver (1.10\%). These results demonstrate SafePath's ability to anticipate collisions more effectively, while operating fully autonomously (HuD = 0\%).


\subsection{Ablation: Impact of LLM2 and Conformal Prediction}\label{app:cp-abl}

The results of the ablation study in Table~\ref{tab:cp-ablation} demonstrate the significant impact of the uncertainty-aware path selection (Stage 2) on improving safety by reducing collision rates. Without conformal prediction (Stages 1+3), the average collision rate is 0.20\%, whereas incorporating Stage 2 reduces it to 0.13\%, a 35\% improvement. This highlights the effectiveness of conformal prediction in pruning high-risk paths, enhancing the reliability of the system.

\begin{table}[ht]
\centering
\scalebox{0.82}{
\begin{tabular}{lllll}
\hline
\multirow{2}{*}{Path Decision} & \multicolumn{4}{c}{Collision (\%)} \\ \cline{2-5}
                               & \multicolumn{1}{c}{1s} & \multicolumn{1}{c}{2s} & \multicolumn{1}{c}{3s} & \multicolumn{1}{c}{Avg} \\ \hline
Stage 1 (no CP)                     & 0.07                   & 0.15                   & 1.10                   & 0.44                    \\
Stages 1+3 (no CP)                  & 0.08                   & 0.18                   & 0.34                   & 0.20                    \\
SafePath (Stages 1+2+3)                       & \cellcolor[HTML]{B9FCD0}\textbf{0.04}                   & \cellcolor[HTML]{B9FCD0}\textbf{0.12}                  & \cellcolor[HTML]{B9FCD0}\textbf{0.24}                   & \cellcolor[HTML]{B9FCD0}\textbf{0.13}                    \\ \hline
\end{tabular}}
\caption{Effects of the conformal prediction and the Uncertainty-Aware Path choice stage on the collision rates at $\alpha=0.3$ and no human intervention ($\delta=0.85$).}
\label{tab:cp-ablation}
\end{table}

\subsection{Closed-Loop Simulation}

\begin{figure}
    \centering
    \includegraphics[width=0.95\linewidth]{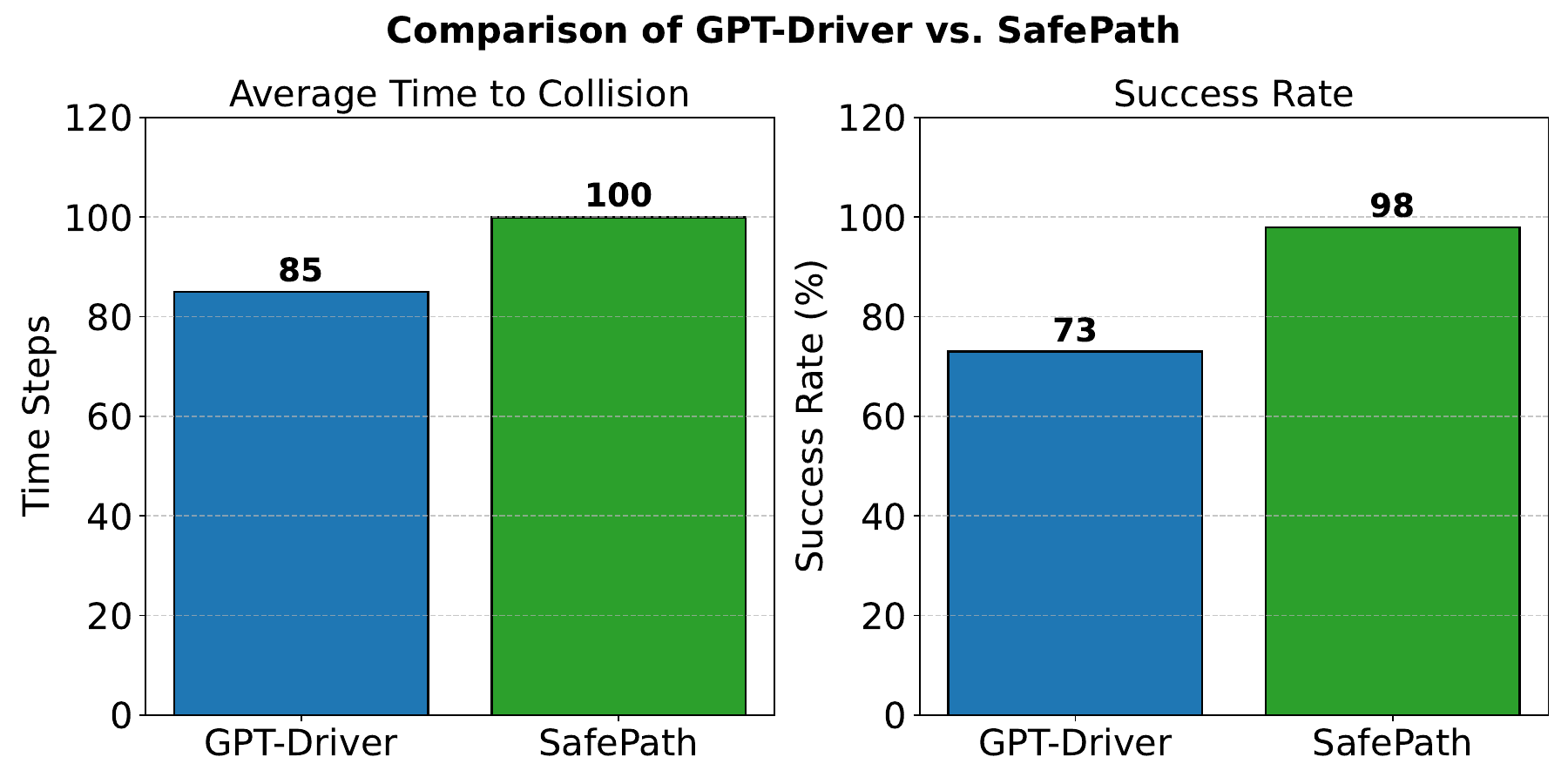}
    \caption{Results of the closed-loop simulation on the Highway-env comparing GPT-Driver to SafePath.}
    \label{fig:closeloopres}
\end{figure}

We evaluate SafePath in a closed-loop setting using the \texttt{highway-env} environments: highway, intersection, and roundabout. Across 10 simulation sequences (100 frames each, 1000 decisions total), we measure average success rate and time to collision, comparing against GPT-Driver~\cite{gpt-driver}. As shown in Figure~\ref{fig:closeloopres}, SafePath achieves a 98\% success rate vs.\ 73\% for GPT-Driver, and a longer time to collision (100 vs.\ 85 frames), indicating more robust autonomous operation. 

\section{Conclusion}
This work introduced SafePath, an approach for safety-centric path planning under uncertainty in autonomous driving, by combining LLMs, conformal prediction, and selective human intervention. SafePath provides formal safety guarantees and demonstrated significant improvements in prediction uncertainty, collision rates, and reduced human intervention compared to state-of-the-art methods.




\section{Limitations}

While SafePath demonstrates strong performance on large-scale datasets, such as nuScenes and simulators such as Highway-env, real-world deployment remains a critical next step. Our results provide a solid foundation, but further validation in controlled environments is essential to evaluate both performance and user acceptance. Controlled experiments, such as those conducted on test tracks or in virtual reality setups~\cite{cyhtt}, can simulate diverse real-world scenarios, enabling a rigorous assessment of SafePath’s applicability and reliability.
The reliance on LLMs for both path generation and selection introduces computational overhead, raising concerns about real-time feasibility. In our current implementation, we relied on APIs, which reduces computational overhead and provides an interface with closed-source LLM such as $\mathtt{gpt-4.o}$, and on lower precision to perform tests with local LLMs, as we show in Appendix~\ref{app:llms}.
Furthermore, the theoretical guarantees provided by CP rely on the exchangeability assumption and the representativeness of the calibration dataset~\cite{cp1, cp2}. While the nuScenes dataset covers challenging scenarios, extending the approach to handle out-of-distribution data (OOD), will enhance its robustness. SafePath is flexible in those terms and is not constrained to a single non-conformity score, making it adaptive to scoring functions that handle OOD~\cite{ood1, ood3}.

\bibliography{main}

\appendix

\clearpage
\setcounter{page}{1}
\startcontents[sections]
\printcontents[sections]{l}{1}{\setcounter{tocdepth}{2}}


\section{Appendix}

\subsection{Proofs}\label{app:proofs}

\subsubsection{Proof for Theorem~\ref{th:safepath_1}}
\begin{proof}
Conformal prediction constructs prediction sets that contain the true output with a specified probability $1 - \alpha$, under the assumption of exchangeability between the calibration data and the test data. In our context, the "true output" refers to a safe path.
By calibrating the confidence scores of the second LLM using a dataset of known safe paths, we ensure that the prediction set $C$ includes the correct (safe) path with probability at least $1 - \alpha$. Therefore, the probability that $C$ contains at least one safe path is at least $1 - \alpha$.
\end{proof}

\subsubsection{Proof for Theorem~\ref{th:safepath_2}}
\begin{proof}
From Theorem \ref{th:safepath_1}, the prediction set $C$ contains at least one safe path with probability at least $1 - \alpha$.

The Path Decision stage proceeds as follows:

\noindent\textbf{Case 1:} If $|C| = 1$, the single path $p$ in $C$ is selected for execution. Since $p$ is in $C$, and $C$ contains at least one safe path with probability at least $1 - \alpha$, the selected path is safe with the same probability.

\noindent\textbf{Case 2:} If $|C| > 1$ and the paths in $C$ are $\delta$-equivalent (i.e., $Sim(p_i, p_j) \geq \delta$ for all $p_i, p_j \in C$), the approach aggregates these paths to produce a final path $p_{\text{final}}$. Given that at least one path in $C$ is safe (from Theorem \ref{th:safepath_1}) and the paths are highly similar, the sampled path $p_{\text{final}}$ will also be safe with probability at least $1 - \alpha$. This is because the sampling from $\delta$-equivalent paths that are close in terms of safety characteristics preserves the safety properties.

\noindent\textbf{Case 3:} If the paths in $C$ are not $\delta$-equivalent, indicating significant divergence among candidate paths or if $C$ is an empty set, the approach delegates the decision to a human operator. Human intervention is assumed to result in a safe action, as the operator can use additional context and judgment not available to the autonomous system.

In all cases, the SafePath approach ensures that the path executed by the AV is safe with probability at least $1 - \alpha$. The combination of statistical guarantees from conformal prediction and the structured decision-making process in the Path Decision stage provides robust safety assurances.

\end{proof}


\subsection{More Results Using L2 Metric} \label{app:l2}

The L2 metric evaluates the distance between the planned and ground-truth trajectories. While L2 distance is commonly used as a performance metric in open-loop predictions, it is important to note that it does not serve as a definitive indicator of the effectiveness of a planning strategy~\cite{l2bad}. This limitation arises from two primary reasons. First, the planned trajectory may differ from the trajectory of the ego-vehicle used to record the dataset, even if both are valid. Second, studies have shown that simple two-layer fully connected neural networks can accurately predict trajectory distributions using only the ego-vehicle's past paths, without incorporating information about the surrounding environment~\cite{l2bad}. This reduces the metric’s ability to meaningfully capture the planner's capability to consider and react to complex external factors. For completeness, we report the results of SafePath on the L2 metric in Table~\ref{tab:sota_l2}. The reported results are obtained using $\alpha$=0.50 and $\delta$=0.90 leading to $HuD= 0\%$.
while maintaining competitive trajectory accuracy. In terms of trajectory accuracy measured by the L2 distance, SafePath has an average L2 error of 0.84 meters. While this is higher than the lowest L2 errors reported by methods such as DriveVLM~\cite{DriveVLM} (0.40m) and VLP-VAD~\cite{vad} (0.50m), it is comparable to GPT-Driver~\cite{gpt-driver} (0.84m) and better than earlier methods like FF~\cite{ff} (1.43m). This suggests that SafePath maintains acceptable trajectory accuracy while emphasizing safety.

\begin{table*}[ht]
\centering
\scalebox{0.95}{
\begin{tabular}{lcccc}
\hline
\multirow{2}{*}{\textbf{Method}} & \multicolumn{4}{c}{\textbf{L2 (m) $\downarrow$}} \\
\cline{2-5}
 & \textbf{1s} & \textbf{2s} & \textbf{3s} & \textbf{Avg.} \\
\hline
NMP~\cite{nmp}       & –    & –    & 2.31 & –    \\
SA-NMP~\cite{nmp}    & –    & –    & 2.05 & –    \\
FF~\cite{ff}         & 0.55 & 1.20 & 2.54 & 1.43 \\
UniAD~\cite{uniad}   & 0.48 & 0.96 & 1.65 & 1.03 \\
VAD-Base (open loop)~\cite{vad}  & 0.41 & 0.70 & 1.05 & 0.72 \\
VLP-VAD (open loop)~\cite{vlp}  & 0.26 & 0.47 & 0.78 & 0.50 \\
DriveVLM (base, open loop)~\cite{DriveVLM}  & \textbf{0.18} &\textbf{0.34} & \textbf{0.68} & \textbf{0.40} \\
GPT-Driver~\cite{gpt-driver}  & 0.27 & 0.74 & 1.52 & 0.84 \\
\hline
SafePath (ours) & 0.31 & 0.53 & 0.79 & 0.54 \\
\hline
\end{tabular}}
\caption{L2 distance (m) performance comparison with deterministic approaches on the nuScenes dataset. For a fair comparison, we compare our model with $0\%$ human delegation ($HuD=0\%$). The best-performing results are highlighted in bold.}
\label{tab:sota_l2}
\end{table*}

\subsection{Effects of Different Path Decision Strategies} \label{app:aggreg}

\begin{table*}[ht]
\centering
\begin{tabular}{lllllllll}
\hline
\multirow{2}{*}{Path Decesion $(\alpha=0.3, \delta=0.85)$} & \multicolumn{4}{c}{L2 (m)}                                                                           & \multicolumn{4}{c}{Collision (\%)}                                                                 \\ \cline{3-4} \cline{7-8}
                                                         & \multicolumn{1}{c}{1s} & \multicolumn{1}{c}{2s} & \multicolumn{1}{c}{3s} & \multicolumn{1}{c|}{Avg}  & \multicolumn{1}{c}{1s} & \multicolumn{1}{c}{2s} & \multicolumn{1}{c}{3s} & \multicolumn{1}{c}{Avg} \\ \hline
Random Sampling                                          & 0.48                   & 0.82                   & 1.26                   & \multicolumn{1}{l|}{0.85} & 0.10                   & 0.34                   & 0.67                   & 0.37                    \\
Average                                                  & \cellcolor[HTML]{B9FCD0}\textbf{0.39}                   & \cellcolor[HTML]{B9FCD0}\textbf{0.68}                   & \cellcolor[HTML]{B9FCD0}\textbf{1.06}                   & \multicolumn{1}{l|}{\cellcolor[HTML]{B9FCD0}\textbf{0.71}} & 0.08                   & 0.24                   & 0.51                   & 0.28                    \\
Conformal Weighted Average                               & \cellcolor[HTML]{B9FCD0}\textbf{0.39}                   & \cellcolor[HTML]{B9FCD0}\textbf{0.68}                   & \cellcolor[HTML]{B9FCD0}\textbf{1.06}                   & \multicolumn{1}{l|}{\cellcolor[HTML]{B9FCD0}\textbf{0.71}} & 0.10                   & 0.24                   & 0.50                   & 0.28                    \\
Conformal Top Path                                       & 0.48                   & 0.82                   & 1.25                   & \multicolumn{1}{l|}{0.85} & 0.12                   & 0.36                   & 0.93                   & 0.47                    \\
minCollision Sampling                                    & 0.48                   & 0.81                   & 1.23                   & \multicolumn{1}{l|}{0.84} & \cellcolor[HTML]{B9FCD0}\textbf{0.08}                   & \cellcolor[HTML]{B9FCD0}\textbf{0.18}                   & \cellcolor[HTML]{B9FCD0}\textbf{0.34}                   & \cellcolor[HTML]{B9FCD0}\textbf{0.20}                    \\ \hline
\end{tabular}
\caption{Different Path Decision strategies at ($\alpha=0.3$, $\delta=0.85$)}
\label{tab:path-decision03}
\end{table*}


\begin{table*}[ht]
\centering
\begin{tabular}{lllllllll}
\hline
\multirow{2}{*}{Path Decesion $(\alpha=0.5, \delta=0.85)$} & \multicolumn{4}{c}{L2 (m)}                                                                                    & \multicolumn{4}{c}{Collision (\%)}                                                                 \\ \cline{3-4} \cline{7-8}
                                                         & \multicolumn{1}{c}{1s} & \multicolumn{1}{c}{2s} & \multicolumn{1}{c}{3s} & \multicolumn{1}{c|}{Avg}           & \multicolumn{1}{c}{1s} & \multicolumn{1}{c}{2s} & \multicolumn{1}{c}{3s} & \multicolumn{1}{c}{Avg} \\ \hline
Random Sampling                                          & 0.47                   & 0.80                   & 1.23                   & \multicolumn{1}{l|}{0.83}          & 0.14                   & 0.32                   & 0.65                   & 0.37                    \\
Average                                                  & \cellcolor[HTML]{B9FCD0}\textbf{0.40}          & \cellcolor[HTML]{B9FCD0}\textbf{0.69}          & \cellcolor[HTML]{B9FCD0}\textbf{1.08}          & \multicolumn{1}{l|}{\cellcolor[HTML]{B9FCD0}\textbf{0.72}} & 0.06                   & 0.16                   & 0.47                   & 0.23                    \\
Conformal Weighted Average                               & \cellcolor[HTML]{B9FCD0}\textbf{0.40}          & \cellcolor[HTML]{B9FCD0}\textbf{0.69}          & \cellcolor[HTML]{B9FCD0}\textbf{1.08}          & \multicolumn{1}{l|}{\cellcolor[HTML]{B9FCD0}\textbf{0.72}} & 0.10                   & 0.18                   & 0.49                   & 0.26                    \\
Conformal Top Path                                       & 0.48                   & 0.81                   & 1.24                   & \multicolumn{1}{l|}{0.84}          & 0.12                   & 0.38                   & 0.73                   & 0.41                    \\
minCollision Sampling                                    & 0.48                   & 0.81                   & 1.24                   & \multicolumn{1}{l|}{0.84}          & \cellcolor[HTML]{B9FCD0}\textbf{0.04}          & \cellcolor[HTML]{B9FCD0}\textbf{0.12}          & \cellcolor[HTML]{B9FCD0}\textbf{0.24}          & \cellcolor[HTML]{B9FCD0}\textbf{0.13}           \\ \hline
\end{tabular}
\caption{Different Path Decision strategies at ($\alpha=0.5$, $\delta=0.85$)}
\label{tab:path-decision05}
\end{table*}

In our experiments, we evaluated the impact of various path decision strategies when the prediction set contains multiple $\delta$-equivalent paths. The strategies examined are:

\begin{enumerate}
    \item \textbf{Random Sampling}: Randomly selecting a path from the prediction set.
    \item \textbf{Average}: Computing the pointwise average of the paths in the prediction set.
    \item \textbf{Weighted Average}: Averaging the paths weighted by their softmax scores.
    \item \textbf{Best Path}: Selecting the path with the highest softmax probability (i.e., lowest non-conformity score).
    \item \textbf{Minimum Collision Sampling}: Choosing the path with the minimum expected collision probability based on input prompts.
\end{enumerate}

Tables~\ref{tab:path-decision03} and \ref{tab:path-decision05} present the results of these strategies at $(\alpha = 0.3, \delta = 0.85)$ and $(\alpha = 0.5, \delta = 0.85)$, respectively, reporting the L2 trajectory error and collision rates at 1\,s, 2\,s, and 3\,s horizons.

The \emph{Average} and \emph{Weighted Average} strategies consistently achieve the lowest L2 trajectory errors across both settings of $\alpha$. Specifically, these strategies yield average L2 errors of 0.71\,m and 0.72\,m at $\alpha = 0.3$ and $\alpha = 0.5$, respectively. This suggests that aggregating paths, whether equally or weighted by confidence scores, leads to more accurate trajectory predictions. The minimal difference between the Average and Weighted Average strategies implies that the softmax confidence scores do not significantly influence the aggregation outcome in terms of trajectory accuracy.

In contrast, the \emph{Minimum Collision Sampling} strategy achieves the lowest collision rates, with average collision rates of 0.20\% at $\alpha = 0.3$ and 0.13\% at $\alpha = 0.5$. This indicates that explicitly selecting the path with the lowest expected collision probability effectively enhances safety. However, this strategy results in higher L2 errors (average of 0.84\,m), suggesting a trade-off between minimizing collision risk and maintaining trajectory accuracy.

The \emph{Random Sampling} and \emph{Best Path} strategies exhibit higher L2 errors and collision rates compared to the averaging strategies and Minimum Collision Sampling. The Best Path strategy, which selects the path with the highest softmax probability, does not perform as well as anticipated. This could be due to calibration issues with the softmax scores or a weak correlation between the confidence scores and actual collision risk. These findings raise concerns about the reliability of using softmax probabilities as indicators of path safety and highlight the need for better calibration of these confidence measures.

These observations highlight a trade-off between trajectory accuracy and collision avoidance. The averaging strategies improve trajectory accuracy by synthesizing information from multiple paths but may not minimize collision risk as effectively as strategies that explicitly consider collision probabilities. Conversely, the Minimum Collision Sampling strategy enhances safety by reducing collisions but may compromise on trajectory precision.

It is important to consider that while averaging paths can produce smoother trajectories, this approach might inadvertently smooth out necessary evasive maneuvers that individual paths would have taken to avoid obstacles, potentially increasing collision risk. Therefore, careful evaluation of how path aggregation impacts safety-critical behaviors is essential.

Overall, the results indicate that no single strategy universally outperforms the others across all metrics. The selection of an appropriate path decision strategy depends on the specific priorities and constraints of the autonomous navigation task. For applications where safety and collision avoidance are highly important, strategies like Minimum Collision Sampling are preferable.


\subsection{Effect of LLM Models on the Conformal Prediction Performance}\label{app:llms}
We evaluate the performance of SafePath with different LLMs at the \emph{Uncertainty Aware Path Selection} stage. LLMs can be biased when it comes to MCQA~\cite{knowno}, which can limit the effects of the conformal prediction operation, especially if the calibration set is limited. So it is crucial to investigate the performance of several LLMs to guide future research.

Table~\ref{tab:llm-cp} presents the results of conformal prediction 3 LLM models: $\mathtt{gpt-4.o}$, $\mathtt{falcon-7B}$, and $\mathtt{llama3.1-7B-instruct}$.
The evaluation of conformal prediction performance across three large language models (LLMs)—$\mathtt{GPT-4.o}$, $\mathtt{falcon-7B}$, and $\mathtt{llama3.1-7B-instruct}$—highlights the interplay between deviation from target coverage (DTC) and prediction set size (SS) under varying significance levels ($\alpha$). These metrics reflect the accuracy and efficiency of uncertainty-aware path planning, critical for autonomous systems operating under uncertainty.

$\mathtt{GPT-4.o}$ consistently achieves the most stable alignment with target coverage, exhibiting minimal DTC values that remain within a small range across $\alpha$ levels. Its ability to maintain compact prediction sets, particularly at higher $\alpha$ levels, underscores its efficiency in balancing coverage and set size. In contrast, $\mathtt{falcon-7B}$ displays variability in DTC, transitioning from under- to over-coverage as $\alpha$ increases, with a fixed SS that indicates limited adaptability. $\mathtt{llama3.1-7B-instruct}$ shows improvement in DTC with increasing $\alpha$, though its larger SS values at lower $\alpha$ levels suggest inefficiencies in stricter confidence scenarios.

Overall, $\mathtt{GPT-4.o}$ emerges as the most robust model, demonstrating strong calibration and efficient prediction set sizing. The observed variability in $\mathtt{falcon-7B}$ and inefficiencies in $\mathtt{llama3.1-7B-instruct}$ highlight opportunities for refining their calibration strategies. These results emphasize the importance of selecting and optimizing LLMs for conformal prediction in applications where accuracy and efficiency are paramount.

\begin{table*}[]
\centering
\begin{tabular}{l|cc|cc|cc|cc|cc}
\hline
                            & \multicolumn{2}{c|}{$\alpha = 0.1$}  & \multicolumn{2}{c|}{$\alpha = 0.2$} & \multicolumn{2}{c|}{$\alpha = 0.3$} & \multicolumn{2}{c|}{$\alpha = 0.4$} & \multicolumn{2}{c}{$\alpha = 0.5$} \\
\multirow{-2}{*}{LLM model} & DTC                           & SS   & DTC               & SS              & DTC                          & SS   & DTC               & SS              & DTC              & SS              \\ \hline
GPT 4.o                     & -0.03                         & 4.82 & -0.03             & 4.62            & -0.05                        & 4.23 & -0.08             & 3.66            & -0.13            & 2.91            \\
Falcon 7B                   & \cellcolor[HTML]{FFFFFF}-0.10 & 4.00 & 0.00              & 4.00            & \cellcolor[HTML]{FFFFFF}0.10 & 4.00 & 0.20              & 4.00            & 0.30             & 4.00            \\
Llama3.1 8B                 & -0.20                         & 4.66 & -0.13             & 4.58            & -0.06                        & 4.44 & 0.01              & 4.29            & 0.06             & 4.29         \\ \hline  
\end{tabular}
\caption{Comparison of different LLM models for conformal prediction. Green cases represent best performing approaches.}
\label{tab:llm-cp}
\end{table*}

\begin{table*}[]
\centering
\begin{tabular}{lllllllll}
\hline
\multirow{2}{*}{\begin{tabular}[c]{@{}l@{}}LLM model\\ ($\alpha= 0.3$, $\delta=0.85$)\end{tabular}} & \multicolumn{4}{c}{L2 (m)}                                                                           & \multicolumn{4}{c}{Collision (\%)}                                                                 \\ \cline{3-4} \cline{7-8}
                                                                                                    & \multicolumn{1}{c}{1s} & \multicolumn{1}{c}{2s} & \multicolumn{1}{c}{3s} & \multicolumn{1}{c|}{Avg}  & \multicolumn{1}{c}{1s} & \multicolumn{1}{c}{2s} & \multicolumn{1}{c}{3s} & \multicolumn{1}{c}{Avg} \\ \hline
GPT 4.o                                                                                             & 0.48                   & 0.81                   & 1.24                   & \multicolumn{1}{l|}{0.83} & 0.04                   & 0.12                   & 0.24                   & 0.13                    \\
Falcon 7B                                                                                           & 0.48                   & 0.82                   & 1.25                   & \multicolumn{1}{l|}{0.85} & 0.06                   & 0.14                   & 0.34                   & 0.18                    \\
Llama3.1 8B                                                                                         & 0.55                   & 0.89                   & 1.30                   & \multicolumn{1}{l|}{0.91} & 0.18                   & 0.34                   & 0.40                   & 0.30                    \\ \hline
\end{tabular}
\caption{Different Path Decision strategies at ($\alpha=0.5$, $\delta=0.85$)}
\label{tab:llm-nu}
\end{table*}

We present results evaluating the performance of three LLMs—$\mathtt{GPT-4.o}$, $\mathtt{falcon-7B}$, and $\mathtt{llama3.1-7B-instruct}$—on the nuScenes dataset for trajectory planning under uncertainty, in Table~\ref{tab:llm-nu}. Using L2 trajectory errors (in meters) and collision rates (in percentages) as evaluation metrics, we assess these models over 1, 2, and 3-second prediction horizons, with an average metric computed across all horizons. The experimental setup corresponds to $\alpha = 0.7$ and $\delta = 0.85$, emphasizing moderately conservative predictions to balance safety and efficiency.

The results show that $\mathtt{GPT-4.o}$ achieves the best overall performance, with an average L2 error of 0.83m and the lowest collision rate of 0.13\%. These metrics indicate its ability to produce precise and safe trajectories across all time horizons. $\mathtt{Falcon-7B}$ performs similarly in terms of L2 error (0.85m average) but has a higher collision rate (0.18\%), suggesting a slight trade-off in safety. $\mathtt{Llama3.1-7B-instruct}$ shows higher L2 error (0.91m average) and collision rates (0.30\%), reflecting weaker performance in trajectory accuracy and safety.

These findings highlight $\mathtt{GPT-4.o}$ as the most effective model for trajectory planning in uncertain environments, excelling in both accuracy and safety. While $\mathtt{falcon-7B}$ and $\mathtt{llama3.1-7B-instruct}$ demonstrate potential, their higher collision rates suggest opportunities for improvement in managing safety-critical scenarios. These results provide valuable insights for optimizing LLM-based planning under uncertainty.


\subsection{Additional Qualitative Results}\label{app:qual1}

\begin{figure*}[ht] 
    \centering
    \begin{subfigure}{\textwidth}
        \includegraphics[width=\textwidth]{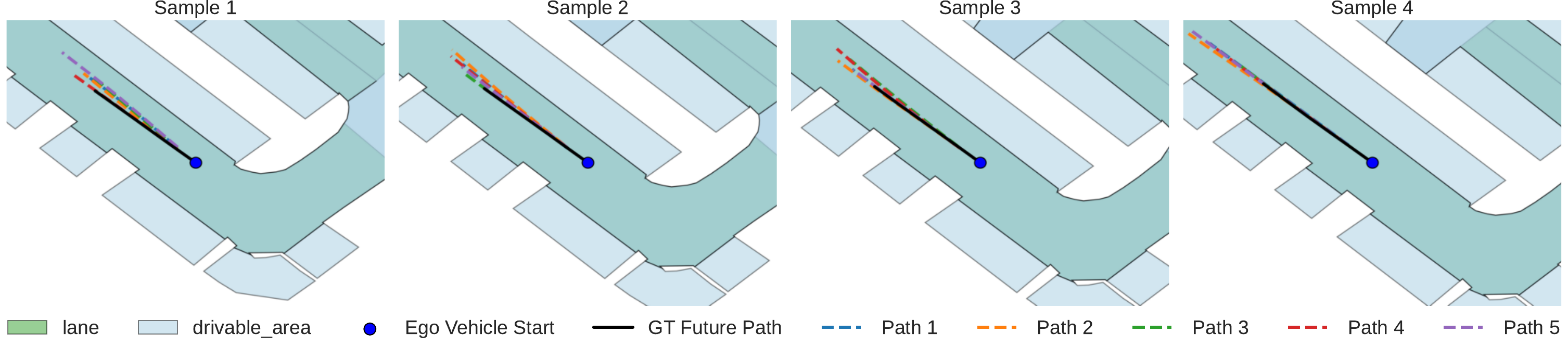}
        \caption{Driving forward -- acceleration}
        \label{fig:qual_subfig1}
    \end{subfigure}
    \hfill 
    \begin{subfigure}{\textwidth}
        \includegraphics[width=\textwidth]{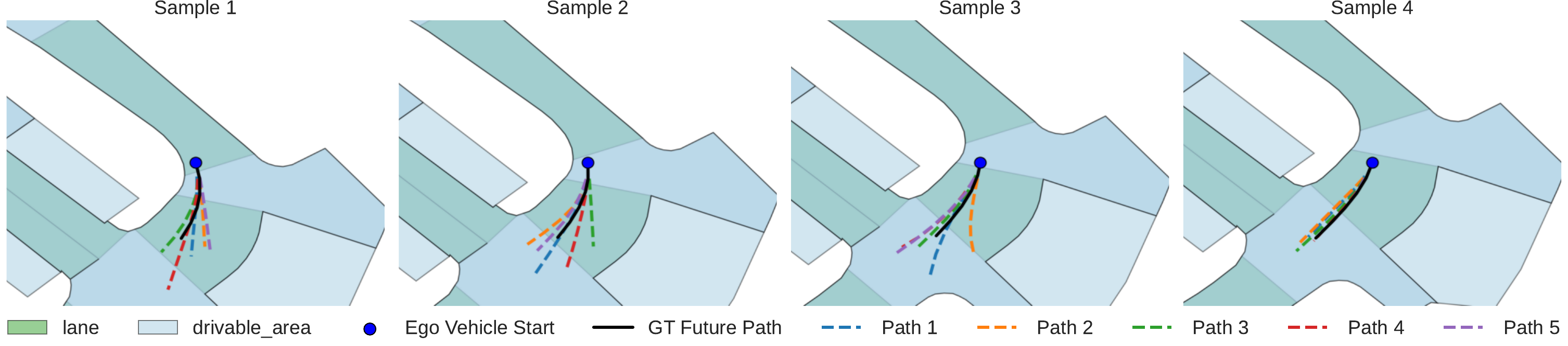}
        \caption{Driving through turn 1}
        \label{fig:qual_subfig2}
    \end{subfigure}
    \hfill 
    \begin{subfigure}{\textwidth}
        \includegraphics[width=\textwidth]{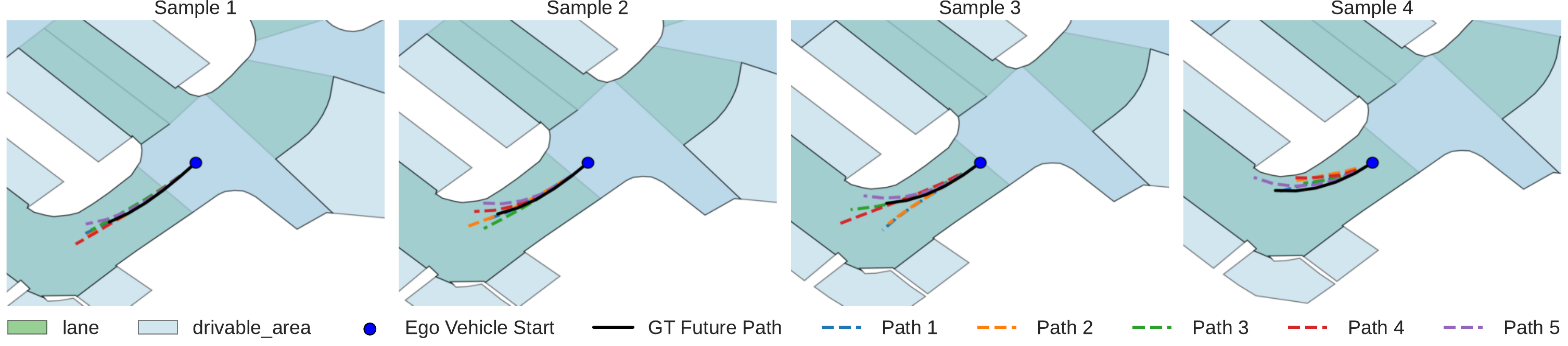}
        \caption{Driving through turn 1}
        \label{fig:qual_subfig3}
    \end{subfigure}

    \caption{Path generation through several successive samples for different situations.}
    \label{fig:qual-matrix}
\end{figure*}

In this section, we present further qualitative results to illustrate SafePath’s performance across various scenarios and configurations. Figure~\ref{fig:qual-matrix} showcases candidate paths generated during the Path Generation stage under diverse situations, including maneuvers such as acceleration (Figure~\ref{fig:qual_subfig1}) and turning scenarios (Figure~\ref{fig:qual_subfig2}). These examples highlight the ability of the approach to propose multiple viable trajectories tailored to the ego vehicle’s goals and the dynamic environmental conditions.

\begin{figure*}[] 
    \centering
    \begin{subfigure}{0.42\textwidth}
        \includegraphics[width=\textwidth]{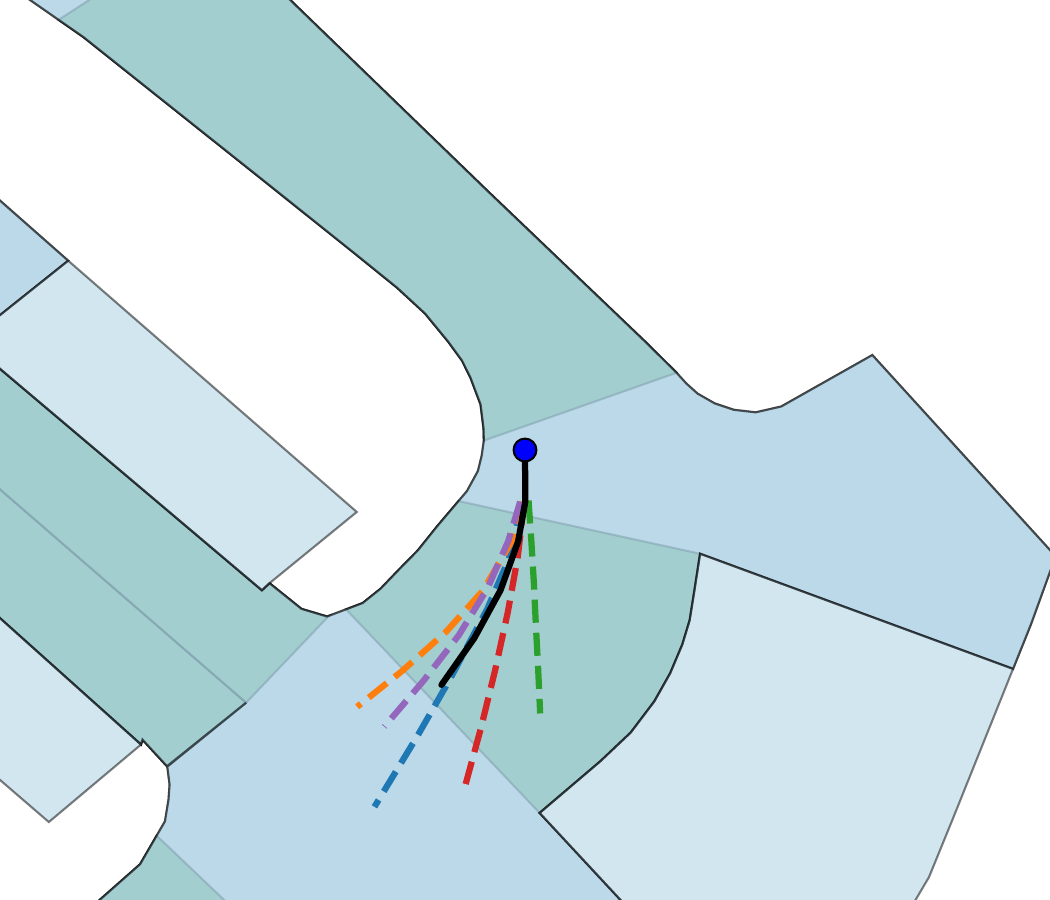}
        \caption{Predicted paths -- situation 1}
        \label{fig:subfig1}
    \end{subfigure}
    \begin{subfigure}{0.42\textwidth}
        \includegraphics[width=\textwidth]{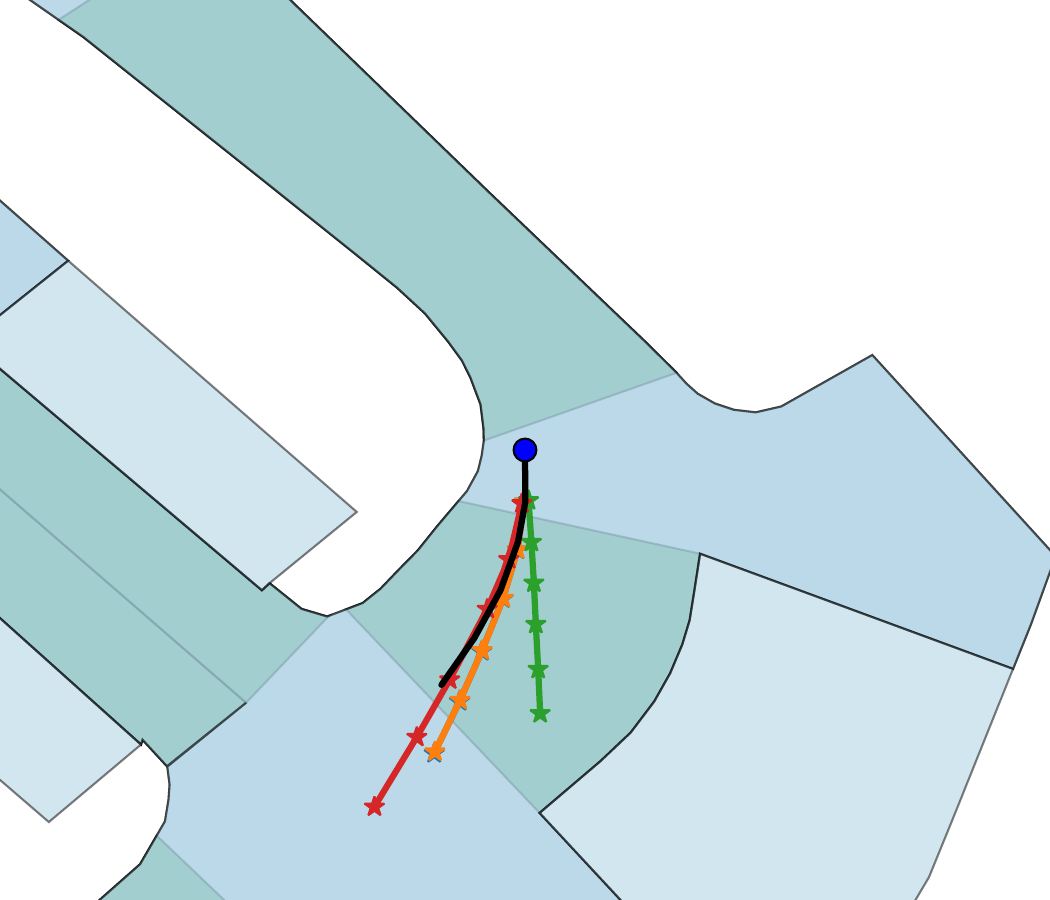}
        \caption{Path decision -- situation 1}
        \label{fig:subfig2}
    \end{subfigure}
    \begin{subfigure}{0.42\textwidth}
        \includegraphics[width=\textwidth]{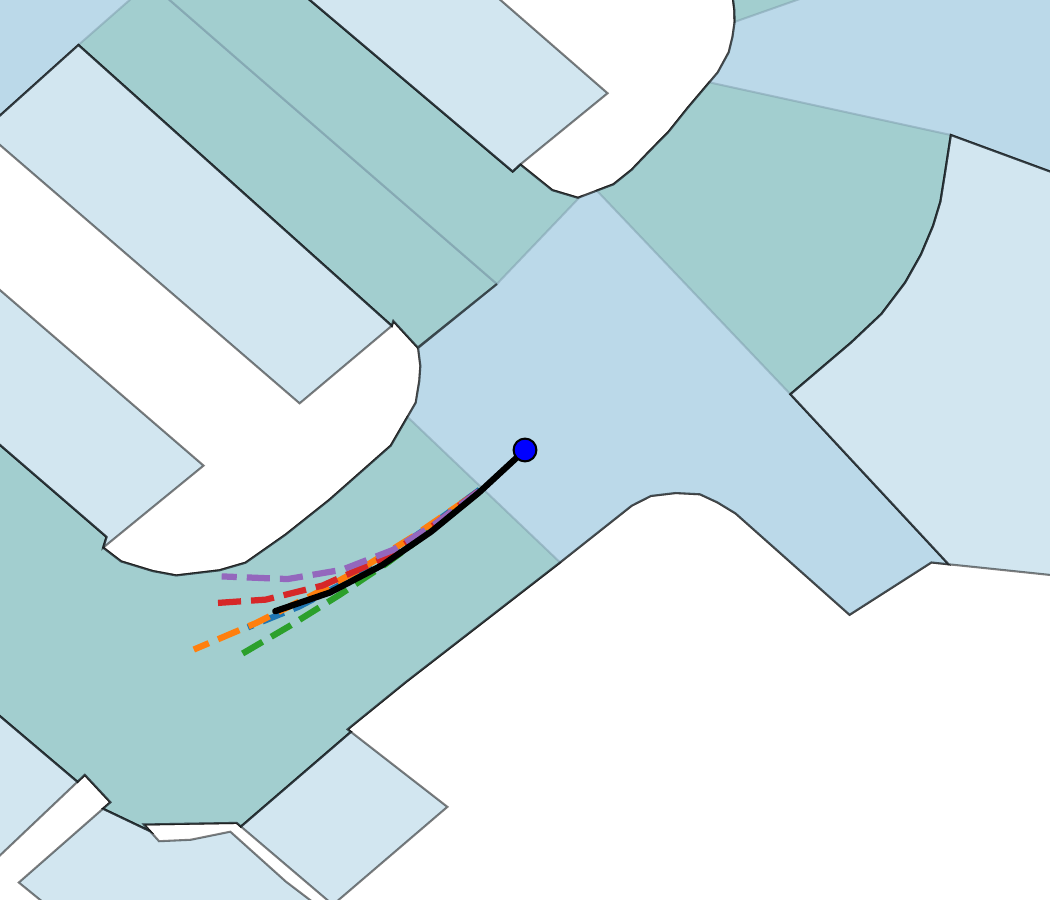}
        \caption{Predicted paths -- situation 2}
        \label{fig:subfig3}
    \end{subfigure}
    \begin{subfigure}{0.42\textwidth}
        \includegraphics[width=\textwidth]{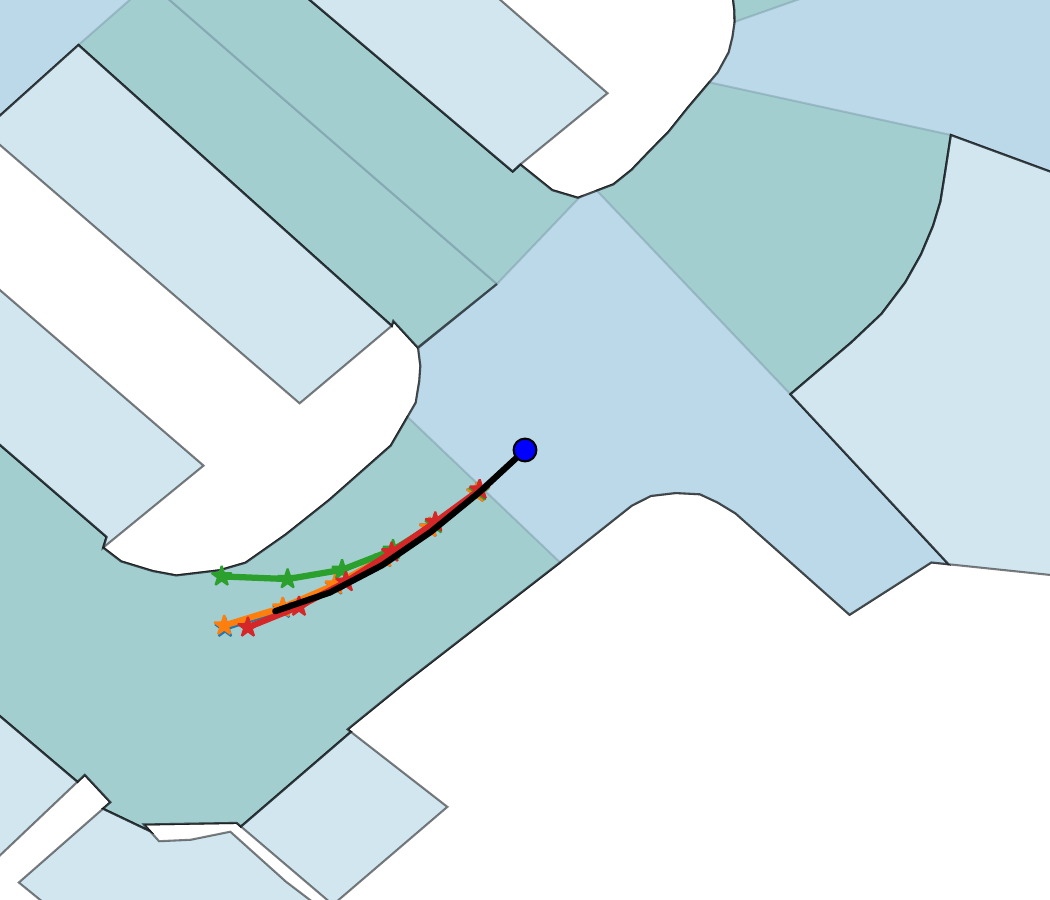}
        \caption{Path decision -- situation 2}
        \label{fig:subfig4}
    \end{subfigure}
    \begin{subfigure}{0.42\textwidth}
        \includegraphics[width=\textwidth]{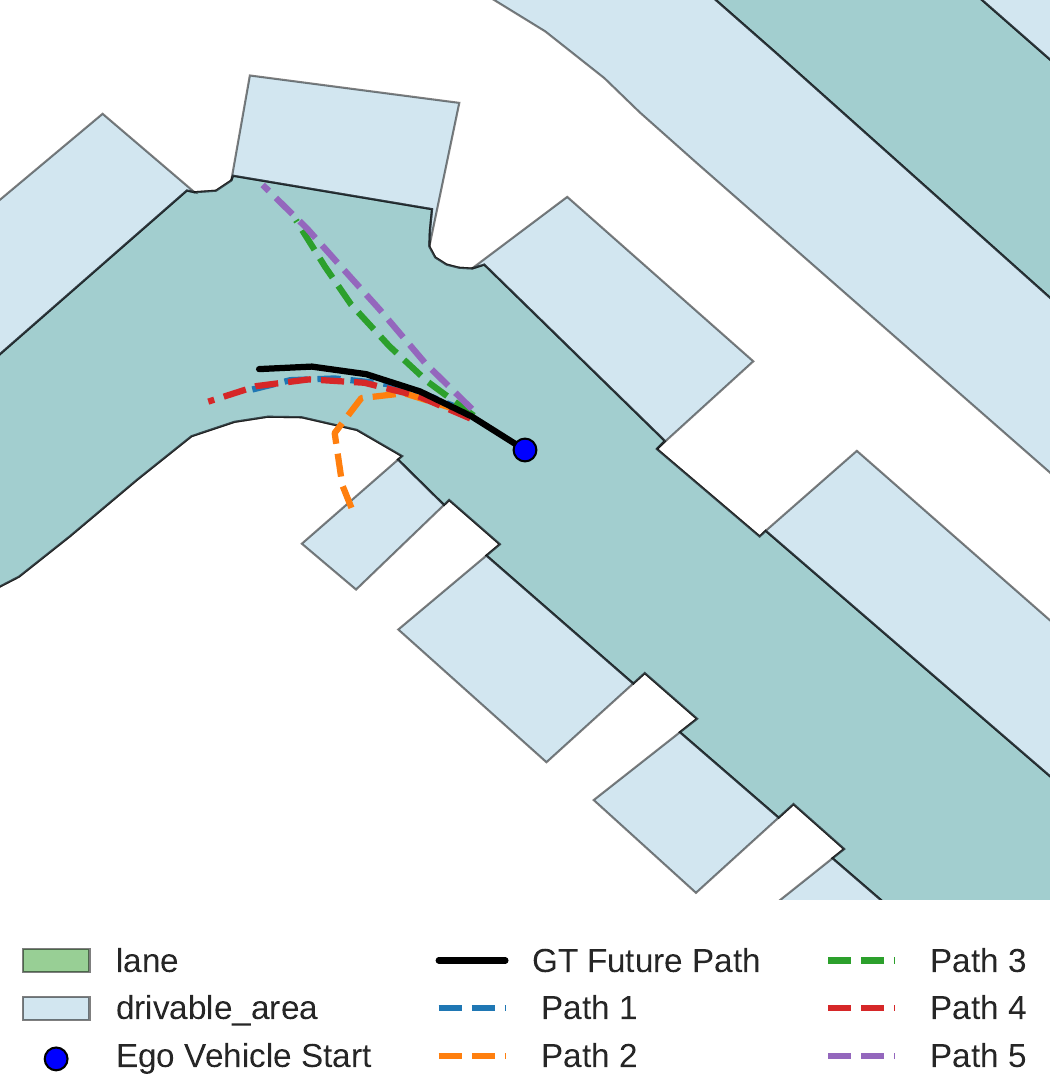}
        \caption{Predicted paths -- situation 3}
        \label{fig:subfig5}
    \end{subfigure}
    \begin{subfigure}{0.42\textwidth}
        \includegraphics[width=\textwidth]{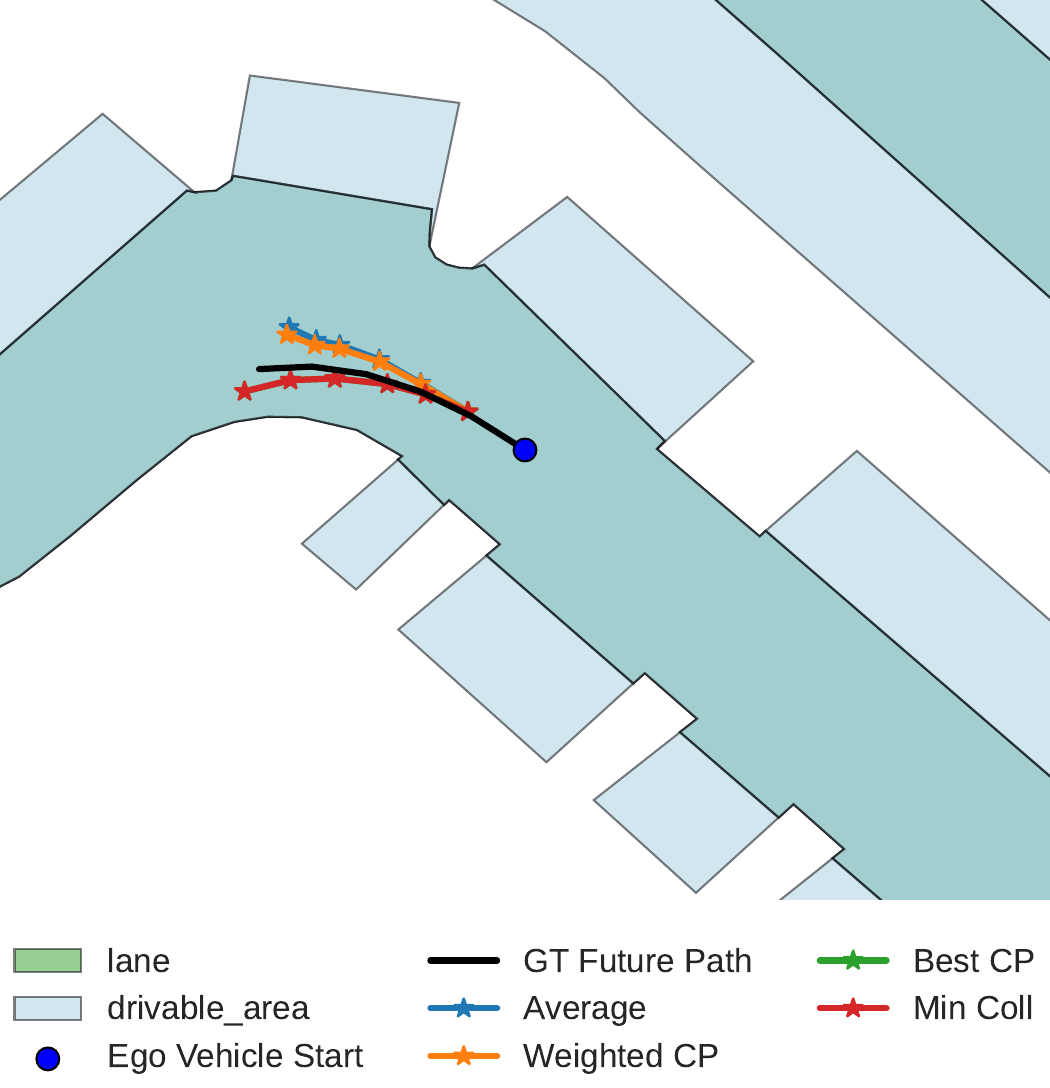}
        \caption{Path decision -- situation 3}
        \label{fig:subfig6}
    \end{subfigure}

    \caption{Predicted paths and path decisions for different situations.}
    \label{fig:qual-matrix-long}
\end{figure*}

Figure~\ref{fig:qual-matrix-long} compares the candidate paths generated in the Path Generation stage with the paths selected using different sampling and path aggregation strategies after constructing the prediction set. Successive frames demonstrate how SafePath refines candidate trajectories to align closely with the ground-truth path while minimizing uncertainty and collision risk. These comparisons underscore the importance of the chosen aggregation strategy, particularly in maintaining trajectory accuracy and safety across dynamic time steps.

Overall, these qualitative results highlight SafePath’s flexibility and robustness in handling diverse navigation scenarios, ensuring safe path selection while adapting to the evolving environment.

\subsection{Prompts}\label{app:prompt}

In the following, we describe the prompts used to fine-tune and query the large language models (LLMs) for trajectory planning and decision-making in the context of autonomous vehicles. Each figure illustrates a specific prompt designed to guide the LLMs through distinct stages of the task, ensuring consistency, safety, and adaptability in path generation and selection under uncertainty.

Figure~\ref{fig:sys-msg} presents the system message used in the trajectory generation phase. The prompt defines the role of the model as a planner tasked with generating four unique, collision-free trajectories over a 3-second horizon. It specifies the inputs, including perception data, historical trajectory, ego-vehicle states, and the mission goal, as well as the expected structure of the output, which includes reasoning, meta-actions, and the planned trajectories.

Figure~\ref{fig:fewshot-msg} demonstrates a few-shot example provided during the trajectory generation process. The example is structured to show how the model processes input data, evaluates the surrounding environment, and produces distinct paths with corresponding thought processes, reasoning, and trajectory plans. The few-shot examples ensure the model adheres to the expected output format and decision rationale.

Figure~\ref{fig:ps-msg} outlines the prompt used for path selection. This stage requires the LLM to evaluate pre-generated trajectory options and select the optimal path based on the mission goal and situational context. The prompt emphasizes concise decision-making, instructing the model to base its choice on reasoning derived from the input data while maintaining alignment with safety and feasibility criteria.

\begin{figure*}[ht]
\centering
\begin{tcolorbox}[colback=black!5!white, colframe=black!80!white, coltext=black, boxrule=0.5mm, arc=2mm, width=0.90\textwidth, title=Autonomous Driving Planner]
\lstset{
    basicstyle=\ttfamily\footnotesize,
    breaklines=true,
    frame=none
}
\begin{lstlisting}
**Autonomous Driving Planner**
Role: You are the brain of an autonomous vehicle. Plan 4 unique, collision-free, safe, and feasible 3-second driving trajectory using 6 new waypoints for each, with each plan including the Thought Process, Reasoning, Meta Action, and Trajectory. Avoid collisions with other objects.

Context
- Coordinates: X-axis is perpendicular, and Y-axis is parallel to the direction you're facing. You're at point (0,0).

Inputs
1. Perception \& Prediction: Info about surrounding objects and their predicted movements.
2. Historical Trajectory: Your past 2-second route, given by 4 waypoints.
3. Ego-States: Your current state including velocity, heading angular velocity, can bus data, heading speed, and steering signal.
4. Mission Goal: Goal location for the next 3 seconds.

Task
- Thought Process: Note down critical objects and potential effects from your perceptions and predictions. Identify immediate risks.
- Reasoning: Provide a deeper analysis of the situation, explaining why certain actions are chosen over others based on predicted outcomes, potential risks, and goal alignment.
- Action Plan: Detail your meta-actions based on your reasoning and thought process.
- Trajectory Planning: Develop 3 unique, safe, and feasible 3-second routes using 6 new waypoints for each, with each plan including the Thought Process, Reasoning, Meta Action, and Trajectory.

Output:
my predicted 4 paths are
Path 1:
- Thoughts:
  - Notable Objects
    Potential Effects
- Reasoning 
- Meta Action
- Trajectories (MOST IMPORTANT):
  - [(x1,y1), (x2,y2), ... , (x6,y6)]

Path 2:
- Thoughts:
  - Notable Objects
    Potential Effects
- Reasoning 
- Meta Action
- Trajectories (MOST IMPORTANT):
  - [(x1,y1), (x2,y2), ... , (x6,y6)]

Path 3:
...

Path 4:
...
\end{lstlisting}
\end{tcolorbox}
\caption{System Message for the path generation stage.}
\label{fig:sys-msg}
\end{figure*}

\begin{figure*}[ht]
\centering
\begin{tcolorbox}[colback=black!5!white, colframe=black!80!white, coltext=black, boxrule=0.5mm, arc=2mm, width=0.9\textwidth, title=Few Shots Example]
\lstset{
    basicstyle=\ttfamily\footnotesize,
    breaklines=true,
    frame=none
}
\begin{lstlisting}

Example 1:
user:
Perception and Prediction:
 - Vehicle at (-4.0, 2.0), stationary.
 - Cyclist at (2.0, 5.0), moving to (2.5, 7.0).
Ego-States:
 - Velocity (vx, vy): (0.00, 1.20)
 - Heading Angular Velocity (v_yaw): 0.05
 - Acceleration (ax, ay): (0.02, -0.01)
Historical Trajectory: [(0.0, -5.0), (0.1, -4.0), (0.2, -3.0), (0.3, -2.0)]
Mission Goal: GO STRAIGHT

assistant:
my predicted 4 paths are
Path 1:
 - Thought Process: Notable Objects: stationary vehicle, and cyclist. Potential Effects: Risk of collision with the cyclist if they intersect our path.
 - Reasoning: The cyclist is moving parallel but could swerve; maintaining awareness is crucial. The stationary vehicle poses no immediate threat.
 - Meta Action: Decelerate slightly to yield to the cyclist, maintain lane position, and proceed cautiously.
 - Trajectory: [(0.0, 1.0), (0.0, 2.0), (0.0, 3.0), (0.0, 4.0), (0.0, 5.0), (0.0, 6.0)]

Path 2:
 - Thought Process: Notable Objects: cyclist. Potential Effects: Shifting lanes could provide additional safety.
 - Reasoning: Moving slightly to the right creates space from the cyclist. Ensures a buffer zone, reducing the risk of collision.
 - Meta Action: Adjust position rightwards while maintaining speed.
 - Trajectory: [(0.5, 1.0), (0.5, 2.0), (0.5, 3.0), (0.5, 4.0), (0.5, 5.0), (0.5, 6.0)]

Path 3:
 - Thought Process: Notable Objects: cyclist. Potential Effects: Stopping allows both to clear the area.
 - Reasoning: A brief stop ensures maximum safety, eliminating movement conflicts. After they pass, proceed to accelerate.
 - Meta Action: Come to a complete stop, then resume driving after clearance.
 - Trajectory: [(0.0, 0.5), (0.0, 1.0), (0.0, 1.5), (0.0, 2.0), (0.0, 4.0), (0.0, 6.0)]

Path 4:
 - Thought Process: Notable Objects: cyclist. Potential Effects: Overtaking the cyclist cautiously.
 - Reasoning: Accelerating slightly to pass the cyclist before potential path crossing. Ensures we're ahead and reduces interaction time.
 - Meta Action: Increase speed moderately to overtake the cyclist safely.
 - Trajectory: [(0.0, 1.5), (0.0, 3.0), (0.0, 4.5), (0.0, 6.0), (0.0, 7.5), (0.0, 9.0)]

\end{lstlisting}
\end{tcolorbox}
\caption{An example from the examples we put in the few shots prompts for the path generation stage. We use a total of 4 examples in our prompts.}
\label{fig:fewshot-msg}
\end{figure*}

\begin{figure*}[ht]
\centering
\begin{tcolorbox}[colback=black!5!white, colframe=black!80!white, coltext=black, boxrule=0.5mm, arc=2mm, width=0.9\textwidth, title=Path Selection Prompt]
\lstset{
    basicstyle=\ttfamily\footnotesize,
    breaklines=true,
    frame=none
}
\begin{lstlisting}

** Path Selection **
You are the decision maker to choose a path that the autonomous vehicle will follow given the Mission Goal. Consider reasoning to make a choice. Based on the following options.

select the best plan based on the following situation:
Perception and Prediction:
 - pushable_pullable at (0.21,12.45).
 - pushable_pullable at (-2.82,1.27).
 - bicycle at (14.43,7.31), moving to unknown location.
 - pushable_pullable at (7.31,10.39), moving to (7.29,10.38).
 - car at (6.25,6.21), moving to (8.39,5.22).
Ego-States:
 - Velocity (vx,vy): (0.00,0.24)
 - Heading Angular Velocity (v_yaw): (-0.00)
 - Acceleration (ax,ay): (0.02,0.25)
 - Can Bus: (2.63,-0.26)
 - Heading Speed: (0.55)
 - Steering: (2.08)
Historical Trajectory (last 2 seconds): [(-0.05,-0.20), (-0.04,-0.21), (-0.04,-0.22), (-0.03,-0.23)]
Mission Goal: RIGHT

** list of paths and reasoning generated by the page generating process (here ommited for space) **

select the best option (A, B, C, D, or E), and respond **only** with the chosen letter, without any additional text.


\end{lstlisting}
\end{tcolorbox}
\caption{Prompt used with LLM2 for the uncertainty-aware path selection process .}
\label{fig:ps-msg}
\end{figure*}

\end{document}